\documentclass[Afour,sageh,times]{sagej}

\usepackage{moreverb,url}

\usepackage[colorlinks,bookmarksopen,bookmarksnumbered,citecolor=red,urlcolor=red]{hyperref}

\usepackage{algorithm}
\usepackage{algpseudocode}
\usepackage{cleveref}
\usepackage{bm}

\newcommand\BibTeX{{\rmfamily B\kern-.05em \textsc{i\kern-.025em b}\kern-.08em
T\kern-.1667em\lower.7ex\hbox{E}\kern-.125emX}}

\newcounter{subeqn} \renewcommand{\thesubeqn}{\theequation\alph{subeqn}}%
\newcommand{\subeqn}{%
	\refstepcounter{subeqn}
	\tag{\thesubeqn}
}
\newcommand{\R}{\mathbb{R}}

\newcommand{\tbx}{\textbf{x}}

\newcommand{\Exp}{\mathbb{E}}

\allowdisplaybreaks

\def\volumeyear{2016}

\begin{document}

\runninghead{Jasour, Han, and Williams}

\title{Convex Risk Bounded Continuous-Time Trajectory Planning and Tube Design in Uncertain Nonconvex Environments}

\author{Ashkan Jasour*, Weiqiao Han*, and Brian Williams}

\affiliation{MIT, Computer Science and Artificial Intelligence Laboratory.\\
* These authors contributed equally to the paper.}

\corrauth{Weiqiao Han, 32 Vassar Street, Rm 235, Cambridge, MA 02139.}
\email{weiqiaoh@mit.edu}

\begin{abstract}
In this paper, we address the trajectory planning problem in uncertain nonconvex static and dynamic environments that contain obstacles with probabilistic location, size, and geometry. To address this problem, we provide a risk bounded trajectory planning method that looks for continuous-time trajectories with guaranteed bounded risk over the planning time horizon. Risk is defined as the probability of collision with uncertain obstacles.
Existing approaches to address risk bounded trajectory planning problems either are limited to Gaussian uncertainties and convex obstacles or rely on sampling-based methods that need uncertainty samples and time discretization. To address the risk bounded trajectory planning problem,
we leverage the notion of risk contours to transform the risk bounded planning problem into
a deterministic optimization problem. Risk contours are the set of all points in the uncertain environment with guaranteed bounded risk. 
The obtained deterministic optimization is, in general, nonlinear and nonconvex time-varying optimization. 
We provide convex methods based on sum-of-squares optimization to efficiently solve the obtained nonconvex time-varying optimization problem and obtain the continuous-time risk bounded trajectories without time discretization. 
The provided approach deals with arbitrary (and known) probabilistic uncertainties, nonconvex and nonlinear, static and dynamic obstacles, and is suitable for online trajectory planning problems.
In addition, we provide convex methods based on sum-of-squares optimization to build the max-sized tube with respect to its parameterization along the trajectory so that any state inside the tube is guaranteed to have bounded risk. 
\end{abstract}


\maketitle

\section{Introduction}\label{sec_intro}
In order for robots to navigate safely in the real world, they need to plan safe trajectories to avoid static and moving obstacles, such as humans and vehicles, under perception uncertainties. The motion planning problem in dynamic environments is known to be computationally hard \cite{reif1994motion}. In this paper, we address the trajectory planning problem in uncertain nonconvex static and dynamic environments that contain obstacles with probabilistic location, size, and geometry. In this problem, the time-varying, nonconvex, and probabilistic nature of the obstacle-free safe regions makes the trajectory planning problem challenging. 

Several approaches have been proposed to address the trajectory planning problems. In the absence of obstacles, one can use standard convex optimization to look for polynomial trajectories that satisfy boundary and way-points conditions \cite{lynch2017modern,traj1, traj2}. In the presence of obstacles, sampling-based methods, including rapidly exploring random tree (RRT) and probabilistic roadmap (PRM), and virtual potential field methods are widely used to find obstacle-free trajectories \cite{lavalle2006planning,lynch2017modern}. In \cite{deits2015efficient}, a mixed-integer optimization is provided for trajectory planning in the presence of convex obstacles. The proposed method first uses convex segmentation to compute convex regions of obstacle-free space. Then, it uses a mixed-integer optimization to assign polynomial trajectories to the computed convex safe regions. Also, \cite{SOS_time2} provides a moment-sum-of-squares-based convex optimization to obtain piece-wise linear trajectories in the presence of deterministic time-varying polynomial obstacles without the need for time discretization.

Trajectory planning problems under uncertainty look for trajectories with a bounded probability of collision with uncertain obstacles. 
Existing methods to address trajectory planning problems under uncertainty either are limited to Gaussian uncertainties and convex obstacles \cite{blackmore2009convex,blackmore2010probabilistic,schwarting2017parallel,luders2010chance,axelrod2018provably,dawson2020provably,dai2019chance} or rely on sampling-based methods \cite{cannon2017chance,calafiore2006scenario,janson2018monte}. 
For example, chance constrained RRT$^*$ algorithm in \cite{luders2010chance} assumes Gaussian uncertainties and linear obstacles and performs a probabilistic collision check for the nodes of the search tree. 
Hence, it can not guarantee to satisfy the probabilistic safety constraints along the edges of the search tree. The Monte Carlo-based motion planning algorithms, e.g., \cite{janson2018monte}, use a large number of uncertainty samples to estimate the probability of collision of a given trajectory. 
Sampling-based methods do not provide any analytical bounds on the probability of collision. 
Instead providing such a certificate usually requires the number of samples to go to infinity.
Due to a large number of samples, sampling-based methods sometimes can be computationally intractable.

Also, \cite{RRT_NONG,wang2020non} use moment-based approaches to address non-Gaussian uncertainties in motion planning problems in the presence of convex obstacles. More precisely, to obtain the risk bounded trajectories, \cite{RRT_NONG} uses first and second-order moments of uncertainties and RRT$^*$ algorithm in the presence of linear obstacles and \cite{wang2020non} uses higher-order moments and interior-point nonlinear optimization solvers in the presence of ellipsoidal obstacles.

\indent\textit{Statement of Contributions}: In this paper, we propose novel convex algorithms for risk bounded continuous-time trajectory planning in uncertain nonconvex static and dynamic environments that contain obstacles with probabilistic location, size, geometry, and trajectories with arbitrary probabilistic distributions. To achieve risk bounded plans:

(1)	We provide an analytical method to compute risk contours maps. Risk contours allow us to identify risk bounded regions in uncertain environments and transform nonlinear stochastic planning problems into deterministic standard planning problems, in the presence of arbitrary (and known) probabilistic uncertainties. Hence, standard deterministic motion planning algorithms, e.g., RRT*, PRM, can be employed to look for safe (risk-bounded) plans.

(2)	To solve the obtained deterministic planning problems, we provide two planners including i) sum-of-squares-based RRT algorithm and ii) sum-of-squares-based convex optimization that allows us to theoretically look for global optimal plans in nonconvex environments.

(3)	We also provide continuous-time safety guarantees in stochastic environments. To ensure safety, existing planning under uncertainty algorithms “only” verify the safety of a finite set of waypoints (time-discretization). Unlike the existing planners, the provided planners of this paper ensure the (risk bounded) safety of the continuous-time trajectories without the need for time discretization.

(4) In addition to planning trajectories, we build tubes around the trajectory so that any states inside the tube are guaranteed to have bounded risk. 
This is achieved via the following two sub-procedures:
\begin{itemize}
    \item[-] Verifying the safety of the tube via convex programming. 
    \item[-] Looking for the max-sized tube with respect to any given parameterization of the tube using binary search.
\end{itemize}

(5) 
We compare our methods with other methods, such as the linear Gaussian method and the Chance-Constrained RRT method.
Specifically, we compare both our analytical method for computing risk contours and our trajectory planning method with the linear Gaussian method. 
And we compare our our tube design method with the Chance-Constrained RRT method.

This article is in continuation of our previous work presented in \cite{jasour2021convex} and the contributions (4) and (5) specifically are the extensions.
The outline of the paper is as follows: Section \ref{sec_def} presents the notation adopted in the paper and definitions of polynomials, moments, and sum-of-squares optimization. In Section \ref{sec_prob}, we provide the problem formulation of risk bounded continuous-time trajectory planning. Section \ref{sec_rc} provides analytical approaches to compute static and dynamic risk contours defined for static and dynamic uncertain obstacles to identify the risk bounded safe regions in uncertain environments. In Section \ref{sec_plan}, using the obtained risk contours, we provide sum-of-squares-based planners to look for continuous-time risk bounded trajectories in uncertain static and dynamic environments. 
In Section \ref{sec_tube}, we extend planning continuous-time risk bounded trajectories to planning continuous-time risk bounded tubes, and we search for the max-sized tube with respect to the given parameterization based on sum-of-squares programming and the binary search algorithm. 
In Section \ref{sec_exp}, we present experimental results on the risk bounded planning problems of autonomous and robotic systems followed by a discussion section. Finally, concluding remarks and future work are given in Section \ref{sec_concl}.

\section{Notations and definitions}\label{sec_def}
This section covers notation and some basic definitions of polynomials, moments of probability distributions, and sum-of-squares optimization.  
For a vector $\mathbf{x}\in \R^n$ and multi-index $\bm{\alpha}\in\mathbb{N}^n$, let $x^{\bm{\alpha}} = \prod_{i=1}^n x_i^{\alpha_i}$. 

\textbf{Polynomials:} Let {$\mathbb{R}[x]$} be the set of real polynomials in the variables { $\mathbf{x} \in \mathbb{R}^n$}. Given polynomial
{$\mathcal{P}(\mathbf{x}):\mathbb{R}^n\rightarrow\mathbb{R}$}, we represent { $\mathcal{P}$} as { $\sum_{\bm{\alpha}\in\mathbb{N}^n} p_{\bm{\alpha}} x^{\bm{\alpha}}$} where 
{ $\{x^{\bm{\alpha}}\}_{\bm{\alpha}\in \mathbb{N}^n}$} are the standard monomial basis of $\mathbb{R}[x]$, { $\mathbf{p}=\{p_{\bm{\alpha}}\}_{\bm{\alpha}\in\mathbb{N}^n}$} denotes the coefficients, and $\bm{\alpha} \in \mathbb N ^n$. In this paper, we use polynomials to describe uncertain obstacles and continuous-time trajectories. For example the set $\{ (x_1,x_2): 1-(x_1-\omega_1)^2-(x_2-\omega_2)^2 \geq 0\}$ represents a circle-shaped obstacle whose center is subjected to uncertainty modeled with random variables $(\omega_1,\omega_2)$. Also, 
$\left[\begin{matrix}
x_1(t)\\
x_2(t)
\end{matrix}\right]=\left[\begin{matrix}
1\\
-1\end{matrix}\right]t + \left[\begin{matrix}
0.5\\
2\end{matrix}\right]t^2$, $t\in [0,1]$ is an example of polynomial trajectory of order 2 in 2D environment between the points $\mathbf{x}(0)=(0,0)$ and $\mathbf{x}(1)=(1.5,1)$.

\textbf{Moments of Probability Distributions:}
Moments of random variables are the generalization of mean and covariance and are defined as
expected values of monomials of random variables. More precisely, given $(\alpha_1,...,\alpha_n) \in \mathbb{N}^n$ where $\alpha=\sum_{i=1}^{n}\alpha_i$, moment of order $\alpha$ of random vector $\mathbf{\omega}$ is defined as $\mathbb{E}[ \Pi_{i=1}^n \omega_i^{\alpha_i}]$. For example, sequence of the moments of order $\alpha=2$ for $n=3$ is defined as 
$\left[\mathbb{E}[w_1^{2}], \mathbb{E}[\omega_1\omega_2],
\mathbb{E}[\omega_1\omega_3],
\mathbb{E}[\omega_2^2] ,
\mathbb{E}[\omega_2\omega_3],
\mathbb{E}[\omega_3^2] \right]$. 
Moments of random variables can be easily computed using the characteristic function of probability distributions \cite{MOM1}. We will use a finite sequence of the moments to represent non-Gaussian probability distributions.

\textbf{Sum of Squares Polynomials and Optimization:} In this paper, we will use sum of squares (SOS) techniques to solve nonconvex optimization problems of the risk bounded trajectory planning problems. Polynomial {$\mathcal{P}(x)$} is a sum of squares polynomial if it can be written as a sum of \emph{finitely} many squared polynomials, i.e., {$\mathcal{P}(x)= \sum_{j=1}^{m} h_j(x)^2$} for some $m<\infty$ and $h_j(x)\in\mathbb{R}[x]$ for $1\leq j\leq m$. 
SOS condition, i.e., {$\mathcal{P}(x) \in SOS$}, can be represented as a convex constraint of the form of a linear matrix inequality (LMI) in terms of the coefficients of the polynomial, i.e.,
{ $\mathcal{P}(x) \in SOS \rightarrow \mathcal{P}(x)=\mathbf{x}^TA\mathbf{x}$} where $\mathbf{x}$ is the vector of standard monomial basis and $A$ is a positive semidefinite matrix in terms of the coefficients of the polynomial \cite{SOS1,SOS2,SOS3}. One can use different software packages like Yalmip \cite{Yalmip_1} and Spotless \cite{Spot_1} to check the SOS condition of the polynomials.

Sum of squares polynomials are used to obtain the convex relaxations of noncovex polynomial optimization problems \cite{SOS1,SOS2,SOS3}. More precisely, consider the following noncovex polynomial optimization as:

 \begin{equation} \label{nlp}
\begin{aligned}
& \underset{\mathbf{x} \in \mathbb{R}^n}{\text{minimize}}
& & f(\mathbf{x}) \\
& \text{subject to}
& & g_i(\mathbf{x}) \geq 0, \ i=1,...,m \\
\end{aligned}
\end{equation}
where $f$ and $g_i, i=1,...,m$ are polynomial functions. We can rewrite the polynomial optimization in \eqref{nlp} as the following form:
\begin{align}\label{nlp_sos}
\underset{\gamma \in \mathbb{R}}{\text{maximize}}\ 
& \gamma\\
\text{subject to}\
& f(\mathbf{x})-\gamma \geq 0, \  \forall \mathbf{x} \in \{\mathbf{x} \in \mathbb{R}^n: g_i(\mathbf{x}) \geq 0 \  |_{i=1}^{m} \}  \nonumber 
\end{align}
where we look for the best lower bound of the function $f(\mathbf{x})$ denoted by $\gamma$ inside the feasible set of the original optimization problem. Hence, if $\mathbf{x}^*$ is the optimal solution of the original optimization problem in \eqref{nlp}, then $\gamma^* = f(\mathbf{x}^*)$ is the optimal solution of the optimization problem in \eqref{nlp_sos}.

Note that in optimization problem \eqref{nlp_sos}, objective function is linear and the constraint is the nonnegativity condition of  polynomial $f(\mathbf{x})-\gamma$. Such nonnegativity condition can be replaced by SOS conditions of polynomials \cite{SOS1,SOS2,SOS3}. Hence, we can transform the optimization in \eqref{nlp_sos} into a convex optimization, i.e., semidefinite program, with LMI constraints in terms of the coefficients of the polynomials of the original optimization in \eqref{nlp}. Also, we can recover the optimal solution of the original polynomial optimization in \eqref{nlp}, i.e., $\mathbf{x}^*$, using the solution of the dual convex optimization problem of \eqref{nlp_sos} as shown in \cite{SOS2,SOS3}. One can use different software packages like GloptiPoly \cite{Glopti} to solve the polynomial optimization in \eqref{nlp} using SOS-based primal-dual approach.

Recently, SOS optimization techniques have been extended to address time-varying polynomial optimization problems of the form
\begin{equation} \label{nlp2}
\begin{aligned}
& \underset{\mathbf{x} \in \mathbb{R}^n}{\text{minimize}}
& & f(\mathbf{x}) \\
& \text{subject to}
& & g_i(\mathbf{x},t) \geq 0,  \ \forall t\in [t_0,t_f], \ i=1,...,m \\
\end{aligned}
\end{equation}
where one needs to make sure that time-varying constraints are satisfied over the given time horizon $t \in [t_0,t_f]$. SOS-based techniques can be used to solve the time-varying optimization problem in \eqref{nlp2} without the need for time discretization by transforming the problem into a convex optimization, i.e., time-varying semidefinite program \cite{SOS_time1,SOS_time2,SOS_time3}.

\section{Problem formulation}\label{sec_prob}
\begin{figure}[H]
    \centering
    \includegraphics[width=0.23\textwidth]{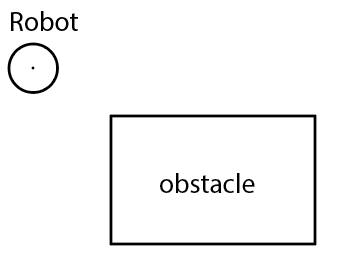}
    \includegraphics[width=0.25\textwidth]{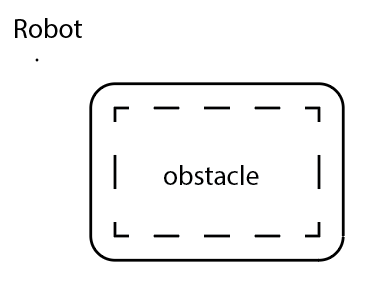}
    \caption{Left: workspace. Right: configuration space.}
    \label{fig:chp1_robot_point}
\end{figure}
We make the following assumptions.
\begin{enumerate}
    \item We mainly consider vehicle models. Most of the time the robot has 2 or 3 degrees of freedom (DOF) and lives in a 2 or 3-dimensional space. We do not consider high degrees of freedom (e.g., 5 DOF, 6 DOF, or 7 DOF) robotic arms.
    \item We assume that the robot is a point. This is a reasonable assumption. Suppose in the 2-dimensional workspace, the robot can be outer-approximated by a disc with radius $r$. Then in the configuration space, the equivalent configuration is that the robot is a point and the obstacle's boundary is extended outwards by a distance of $r$ (\Cref{fig:chp1_robot_point}). 
    The 3-dimensional case is analogous. 
    A state space usually refers to a configuration space, though sometimes can also refer to a workspace. 
    \item In addition to planning trajectories, we also plan tubes, which is a 2-dimensional region in 2-dimensional state space, or a 3-dimensional solid in 3-dimensional state space. 
    One of the motivations of planning tubes is that we work directly in the workspace where the robot is outer-approximated by a disc in 2-dimensional state space or a ball in 3-dimensional state space.
    \item We assume the uncertainty appeared in any problem does not necessarily have a Gaussian distribution. However, we assume that the moments of the probability distribution can be calculated and are finite.
    \item We assume obstacles are represented by polynomials. 
    They can be of arbitrary shape, can deform over time, can move, and have arbitrary uncertainty. 
    The change in shapes and positions can be captured by polynomials parameterized by time $t$, and the uncertainty can be represented by random variables in the polynomial. 
    In essence, we are considering the trajectory planning problem in the presence of ``known unknowns". 
    This is a reasonable assumption, because there are many techniques that predict the future trajectories of obstacles.
    For example, in the recently developed autonomous driving research community, there is a whole area devoted to predicting the future trajectories of agents, including vehicles, bicyclists, and pedestrians, in the traffic scene \cite{gu2021densetnt, lu2022kemp}. 
    After getting information from the environment via perception, the robot first predicts the future trajectories of other agents in the scene and then plans its own trajectory.
    The predicted trajectories can be Gaussian mixture models (GMM) \cite{chai2019multipath,varadarajan2022multipath++}, and they satisfy our assumption.
\end{enumerate}

Suppose $\mathcal{X}\in \mathbb{R}^{n_x}$ is an uncertain environment and the sets $\mathcal{X}_{obs_i}(\omega_i) \subset \mathcal{X}, \ i=1,...,n_{o_s}$ and
$\mathcal{X}_{obs_i}(\omega_i,t) \subset \mathcal{X}, \ i=n_{o_s}+1,...,n_{o_d}$ are 
the static and dynamic uncertain obstacles, respectively, where $\omega_i \in \mathbb{R}^{n_{\omega}}, \ i=1,...,n_{o_s}+n_{o_d}$
are probabilistic uncertain parameters with known probability distributions. We represent static uncertain obstacles in terms of polynomials in $\mathbf{x}\in \mathcal{X}$ and uncertain parameters as follows:
\begin{equation} \label{obs_s}
\mathcal{X}_{obs_i}(\omega_i)= \{ \mathbf{x}\in \mathcal{X}: \mathcal{P}_i(\mathbf{x},\mathbf{\omega}_i) \geq 0  \}, \ i=1,...,{n_{o_s}}
\end{equation}
where $\mathcal{P}_i:  \mathbb{R}^{n_x+n_{\omega}} \rightarrow \mathbb{R}, i=1,...,n_{o_s}$ are the given polynomials. Similarly, we represent dynamic uncertain obstacles in terms of polynomials in $\mathbf{x}\in \mathcal{X}$, time $t$, and uncertain parameters as follows:
\begin{equation} \label{obs_d}
\mathcal{X}_{obs_i}(\omega_i,t)= \{ \mathbf{x}\in \mathcal{X}: \mathcal{P}_i(\mathbf{x},\mathbf{\omega}_i, t) \geq 0  \} \ |_{i=n_{o_s}+1}^{n_{o_d}}    
\end{equation}
where $\mathcal{P}_i:  \mathbb{R}^{n_x+n_{\omega}+1} \rightarrow \mathbb{R}, i=n_{o_s}+1,...,n_{o_d}$ are the given polynomials. Note that, in
general, the sets in \eqref{obs_s} and \eqref{obs_d} represent \textit{nonconvex} probabilistic obstacles, e.g., nonconvex obstacles with uncertain size, location, or geometry \cite{Contour,Risk_Ind}.

\textbf{Continuous-time trajectory planning:}
Given static and dynamic uncertain obstacles in \eqref{obs_s} and \eqref{obs_d}, we define risk as the probability of collision with uncertain obstacles in the environment. In the risk bounded continuous-time trajectory planning problem, we aim at finding a continuous-time trajectory $\mathbf{x}(t): [t_0, t_f] \rightarrow \mathbb{R}^{n_x}$ defined over the time horizon $t\in[t_0,t_f]$ between the start and final points $\mathbf{x}_0$ and $\mathbf{x}_f$ such that the probability of collision of the trajectory $\mathbf{x}(t)$ with uncertain obstacles is bounded. 

More precisely, we define the risk bounded continuous-time trajectory planning problem as the following probabilistic optimization problem:
\begin{align} 
& \underset{\mathbf{x}(t): [t_0, t_f] \rightarrow \mathbb{R}^{n_x}}{\text{minimize}}
 \int_{t_0}^{t_f}  \left\Vert \dot{\mathbf{x}}(t) \right\Vert_2^2 dt \label{opt_obj} \\
& \text{subject to}  \ 
\ \ \mathbf{x}(t_0)=\mathbf{x}_0, \  \mathbf{x}(t_f)=\mathbf{x}_f \subeqn \\
&  \ \hbox{Prob}\left( \mathbf{x}(t) \in \mathcal{X}_{obs_i}(\omega_i)    \right) \leq \Delta, \ \forall t\in [t_0,t_f] \ |_{i=1}^{n_{o_s}} \subeqn \label{opt_prob1} \\
& \  \hbox{Prob}\left(  \mathbf{x}(t) \in \mathcal{X}_{obs_i}(\omega_i,t)    \right) \leq \Delta, \ \forall t\in [t_0,t_f] \ |_{i=n_{o_s}+1}^{n_{o_d}} \subeqn \label{opt_prob2}
\end{align}
\noindent where objective function \eqref{opt_obj} tries to minimize the length of the trajectory. Also, constraints \eqref{opt_prob1} and \eqref{opt_prob2} are the defined risks at time $t$ for trajectory $\mathbf{x}(t)$ in terms of the uncertain static and dynamic obstacles, respectively. Moreover, $ 0 \leq \Delta \in \mathbb{R} \leq 1 $ is the given acceptable risk level.
To solve the risk bounded optimization problem in \eqref{opt_obj}, we will look for the following continuous-time trajectories:

i) Polynomial trajectories over the planning horizon $[t_0,t_f]$ of the form 
\begin{equation}\label{poly}
\mathbf{x}(t)=\sum_{\alpha=0}^d \mathbf{c}_{\alpha}t^{\alpha}, \  t\in[t_0,t_f]    
\end{equation}
where $\mathbf{c}_{\alpha} \in \mathbb{R}^{n_x}, \ \alpha=0,...,d$ are the coefficient vectors,

ii) Piecewise linear trajectories of the form
\begin{equation}\label{linear}
    \mathbf{x}_i(t)= \mathbf{a}_i+ \mathbf{b}_it, \ t \in [t_{i-1}, t_i), \ i=1,...,s
\end{equation}
where $s$ is the number of linear pieces defined over the time intervals $t \in [t_{i-1},t_i] $ of the form $t_{i-1}=t_0+\frac{(i-1)(t_f-t_0)}{s}$ and $t_{i} = t_0+ \frac{i(t_f-t_0)}{s}, i=1,...,s$, and $\mathbf{a}_i,\mathbf{b}_i \in \mathbb{R}^{n_x}$ are the coefficient vectors.

Solving the probabilistic optimization in \eqref{opt_obj} is challenging,  because i) we need to deal with multivariate integrals of the probabilistic constraints in \eqref{opt_prob1} and \eqref{opt_prob2} defined over the nonconvex sets of the obstacles, ii)
we need to deal with time-varying constraints to ensure that they are all satisfied over the entire planning time horizon $[t_0,t_f]$, and iii) optimization in \eqref{opt_obj} is, in general, nonconvex optimization; Hence, we cannot guarantee to obtain the global optimal solution.

\textbf{Continuous-time tube design:}
A tube is an $n$-dimensional region in the $n$-dimensional state space at any time.
For example, while a trajectory is a 1-dimensional point moving in the 2-dimensional or 3-dimensional state space over time, a tube can be a 2-dimensional disc moving in the 2-dimensional state space or a 3-dimensional solid moving in the 3-dimensional state space over time. 
We build a tube around a risk bounded trajectory so that any states inside the tube are guaranteed to have bounded risk. 
The reasons for building tubes are three-folds:
\begin{enumerate}
    \item When we work directly in the workspace (\Cref{fig:chp1_robot_point} Left), the robot can be outer-approximated by a disc in the 2-dimensional state space, or by a ball in the 3-dimensional state space. We need to build tubes, which is an $n$-dimensional region at any time in the $n$-dimensional state space, instead of a point at any time as in the trajectory case, to ensure the safety of the robot. The nominal trajectory of the tube can be viewed as the trajectory of the center of the robot.
    \item When we work in the configuration space (\Cref{fig:chp1_robot_point} Right), the tube represents a collection of trajectories that are safe. If the robot is able to follow any trajectory exactly without any deviations, then the robot can follow any trajectory in the tube to reach the goal without violating any safety constraints. 
    \item Once we consider the dynamics of the robot, the planned continuous time trajectory can be viewed as a high level guidance for the robot to follow. Due to uncertainty in the dynamics, the robot can deviate from the nominal trajectory. However, as long as the robot stays in the tube, it is guaranteed to be safe. 
\end{enumerate}
A tube can be designed in many ways. 
For example, at any time, a tube can be an $n$-dimensional ball or an $n$-dimensional cube in the $n$-dimensional state space. 
We call such a design choice a tube parameterization. 
We define a tube as states near a planned trajectory $\bar{\tbx}(t)$:
\begin{equation}\label{tube}
\mathcal{Q}(\bar{\tbx}(t),t) =\{ \mathbf{x} \in \mathbb{R}^{n_x}:  (\mathbf{x}-\bar{\tbx}(t))^T{Q(t)}(\mathbf{x}-\bar{\tbx}(t))\leq 1 \}   
\end{equation}
where $Q(t)$ is a positive-definite matrix, for any $t \in [t_0,t_f]$.
For example, if $Q(t)$ is a multiple of the identity matrix, then the tube is a ball centered at $\bar{\tbx}(t)$ for any $t$.
In the risk bounded continuous-time tube design problem, we aim to find a tube $\mathcal{Q}(\bar{\tbx}(t),t)$ associated with a continuous-time trajectory $\bar{\tbx}(t)$ over the time horizon $[t_0,t_f]$ starting from $\tbx_0$ and ending at $\tbx_f$, such that the probability of collision of any state in the tube $\mathcal{Q}(\bar{\tbx}(t),t)$ with uncertain obstacles is bounded, and we aim to maximize the size of such a tube with respect to the parameterization of the tube. 

More precisely, we define the risk bounded continuous-time tube design problem as the following probabilistic optimization problem:
\begin{align} 
& \underset{\substack{\mathbf{x}(t): [t_0, t_f] \rightarrow \mathbb{R}^{n_x}\\ Q(t) \succ 0, t\in [t_0, t_f]}}{\text{minimize}}
 \int_{t_0}^{t_f}  \left\Vert \dot{\mathbf{x}}(t) \right\Vert_2^2 dt + w\max_{t\in[t_0,t_f]} \lambda_{max}(Q(t))\label{opt2_obj} \\
& \text{subject to}  \ 
\ \ \mathbf{x}(t_0)=\mathbf{x}_0, \  \mathbf{x}(t_f)=\mathbf{x}_f \tag{\ref{opt2_obj}a} \\
&  \ \hbox{Prob}\left( \mathcal{Q}({\tbx}(t),t) \cap \mathcal{X}_{obs_i}(\omega_i) \neq \emptyset    \right) \leq \Delta, \ \forall t\in [t_0,t_f] \ |_{i=1}^{n_{o_s}} \tag{\ref{opt2_obj}b} \label{opt2_prob1} \\
& \  \hbox{Prob}\left(  \mathcal{Q}({\tbx}(t),t) \cap \mathcal{X}_{obs_i}(\omega_i,t)    \neq \emptyset \right) \leq \Delta, \nonumber \\ 
& \quad\quad\quad\quad\quad\quad\quad\quad\quad\quad\quad\quad \forall t\in [t_0,t_f] \ |_{i=n_{o_s}+1}^{n_{o_d}} \tag{\ref{opt2_obj}c} \label{opt2_prob2}\\
& \ \mathcal{Q}({\tbx}(t),t) \subseteq M(t), \forall t\in [t_0,t_f] \tag{\ref{opt2_obj}d} \label{opt2_constr4}
\end{align}
where $\lambda_{max}(Q(t))$ represents the largest eigenvalue of $Q(t)$ and the second term in the objective is the maximum of all maximum eigenvalues of $Q(t)$, for each time, over the entire time horizon, weighted by a hyperparameter $w$. 
By minimizing the maximum eigenvalues of $Q(t)$, we are trying to maximize the size of the tube. 
For example, if $Q(t)$ is a multiple of an identity matrix for any $t$, then the tube is a ball for any $t$, and the eigenvalues of $Q(t)$ are the inverse of squared radius of the ball for any $t$.
In this case, minimizing the maximum eigenvalues of $Q(t)$ amounts to maximizing the radius of the ball.
The first term in the objective tries to minimize the trajectory length.
The weight $w$ is meant to balance maximizing the tube size and minimizing the trajectory length. 
Constraints \eqref{opt2_prob1} and \eqref{opt2_prob2} are the risks at time $t$ for the tube $\mathcal{Q}({\tbx}(t),t)$ in terms of the uncertain static and dynamic obstacles, respectively, and $ 0 \leq \Delta \in \mathbb{R} \leq 1 $ is the given acceptable risk level. 
The constraint \eqref{opt2_constr4} imposes an upper bound on the size of the tube. 
For example, if the tube is a ball for any $t$, then we impose an upper bound on the radius of the ball.
This is reasonable, because for example we do not want the tube to grow outside the state space.
While we consider polynomial and piecewise linear trajectories in trajectory planning, we consider the union of balls around the trajectory as a tube, in the form of
\begin{align}\label{eq:tube_form}
    \mathcal{Q}(\bar{\tbx}(t),t) = \{\tbx \in \R^{n_x}: r(t)^2 - ||\tbx - \bar{\tbx}(t)||_2^2 \geq 0\}
\end{align}
So the constraint \eqref{opt2_constr4} translates to an upper bound on the radius $r(t)$.
The probabilistic optimization \eqref{opt2_obj} is even more challenging than \eqref{opt_obj}, because it is nonconvex and has additional terms involving the tube. The risk bound needs to hold for a whole region instead of a single trajectory.

In this paper, we provide a systematic numerical procedure to efficiently solve the probabilistic optimization problems in \eqref{opt_obj} and \eqref{opt2_obj} in the presence of nonconvex uncertain obstacles with arbitrary probability distributions. For this purpose, we will leverage the notion of risk contours to transform the probabilistic optimization in \eqref{opt_obj} and \eqref{opt2_obj} into deterministic polynomial optimization problems.
We will use SOS optimization techniques to obtain optimal continuous-time trajectories with guaranteed bounded risk. 
We will also use SOS optimization and binary search to look for max-sized tube with guaranteed bounded risk.

\section{Risk contours} \label{sec_rc}
In \cite{Contour}, we define the risk contour with respect to the static uncertain obstacle in \eqref{obs_s} and the given acceptable risk level
$\Delta$ in \eqref{opt_prob1} as the set of all points in the environment whose probability of collision with the uncertain obstacle is less or equal to $\Delta$.
In this paper, we use static and dynamic risk contours defined for static and dynamic uncertain obstacles, respectively, to identify the safe regions in uncertain environments, i.e., the feasible set of optimization \eqref{opt_obj}.

In \cite{Contour}, to construct the risk contours of uncertain static obstacles, we propose an $(n_x+n_{\omega})$-dimensional convex optimization in the form of a semidefinite program (SDP). Such optimization is not suitable for online computations and is limited to small dimensions $(n_x+n_{\omega})$. In this paper, we propose an optimization-free fast approach, i.e., an analytical method, to construct the risk contours both for static and dynamic uncertain obstacles and show how one can use the obtained risk contours to solve the risk-bounded trajectory planning problem in \eqref{opt_obj}.

\subsection{Static Risk Contours} \label{sec_rc_s}
Let $\mathcal{X}_{obs}(\omega)= \{ \mathbf{x}\in \mathcal{X}: \mathcal{P}(\mathbf{x},\mathbf{\omega}) \geq 0  \}$ be the given static uncertain obstacle as defined in \eqref{obs_s} and $\Delta \in [0,1]$ be the given acceptable risk level. Then, static $\Delta$-risk contour denoted by $\mathcal{C}^{\Delta}_{r}$ is defined as the set of all points in the environment, i.e., $\mathbf{x} \in \mathcal{X}$, whose risk is less or equal to $\Delta$. More precisely,
\begin{equation}\label{rc_s}
\mathcal{C}^{\Delta}_{r}:= \{\ \mathbf{x} \in \mathcal{X}: \ \hbox{Prob}( \mathbf{x} \in \mathcal{X}_{obs}(\omega)) \leq \Delta  \}
\end{equation}

The main idea to construct the static risk contour in \eqref{rc_s} is to replace the probabilistic constraint, i.e., 
$\hbox{Prob}( \mathbf{x} \in \mathcal{X}_{obs}(\omega))=\hbox{Prob}( \mathcal{P}(\mathbf{x},\omega)\geq 0 ) \leq \Delta$, with a deterministic constraint in terms of $\mathbf{x}$. 
We are going to propose an analytical method to compute risk contours. For comparison purposes, we first introduce an optimization-based method, and show that it is inefficient and can be more conservative compared to our method.

\subsubsection{Optimization-based Method}
The optimization-based method tries to find a polynomial upper bound of the indicator function associated with the risk. 
The problem of finding a polynomial upper bound can be formulated as an SOS programming problem. 

To compute the static risk contour in \eqref{rc_s}, we obtain the deterministic constraint as follows:
For a given point $\mathbf{x} \in \mathcal{X}$, the probability of collision with the uncertain static obstacle, i.e., $\hbox{Prob}( \mathbf{x} \in \mathcal{O}(\omega))$, is equivalent to the expectation of the indicator function of the superlevel set of $p(\mathbf{x},\omega)$ as follows \cite{Contour,Risk_Ind}:

\begin{equation}
\hbox{Prob}( p(\mathbf{x},\omega)\geq 0 ) = \int_{ \{ \mathbf{\omega}: p(\mathbf{x},\omega)\geq 0\}} pr(\mathbf{\omega})d\mathbf{\omega}= \mathbb{E}[\mathbb{I}_{p\geq 0}]
\end{equation}
where $pr(\mathbf{\omega})$ is the probability density function of $\mathbf{\omega}$ and $\mathbb{I}_{p\geq 0}$ is the indicator function of the superlevel set of $p(\mathbf{x},\omega)$ defined as $\mathbb{I}_{p\geq 0} = 1 $ if $(\mathbf{x},\omega) \in  \{ (\mathbf{x},\omega): p(\mathbf{x},\omega) \geq 0\} $, and
0 otherwise. The expectation of the indicator function, however, is not necessarily easily computable. To compute the expectation value, we want to find a polynomial description of the indicator function that upper bounds the true indicator function $\mathbb{I}_{p\geq 0}$. If we can find a polynomial $P_{\mathbb{I}}: \R^{n_x+n_{\omega}}\rightarrow\R$ of the order $d$ with coefficients $c_{ij}$ that upper bounds the indicator function, i.e., $  P_{\mathbb{I}}(\mathbf{x},\omega):= \sum_{(i,j)} c_{ij} x^{i} \omega^{j} \geq \mathbb{I}_{p \geq 0}$, 
then we can apply the expectation w.r.t. the probability density function of $\mathbf{\omega}$ to the both sides and describe the upper bound probability in terms of the known moments of $\mathbf{\omega}$ as follows:
\begin{align}\label{risk_ind}
\mathbb{E}[P_{\mathbb{I}}(\mathbf{x},\omega)]&=\sum_{(i,j)} c_{ij} x^{i} \mathbb{E}[\omega^{j}]  \nonumber\\
&\geq \mathbb{E}[\mathbb{I}_{p \geq 0}]
=\hbox{Prob}( p(\mathbf{x},\omega)\geq 0 ) 
\end{align}
where $\mathbb{E}[\omega^j]$ is the moment of order $j$ of random vector $\omega$. 
Hence, we can construct an inner approximation of the $\Delta$-risk contour in \eqref{rc_s}, denoted by $\hat{\mathcal{C}}^{\Delta}_{r}$, using the upper bound probability in \eqref{risk_ind} as follows:

\begin{equation}\label{rc_s_inner}\hat{\mathcal{C}}^{\Delta}_{r}= \{\ \mathbf{x} \in \mathcal{X}: \ P(\mathbf{x})
\leq \Delta  \}
\end{equation}
where polynomial $P(\mathbf{x})=\mathbb{E}[P_{\mathbb{I}}(\mathbf{x},\omega)]=\sum_{(i,j)} c_{ij} \mathbb{E}[\omega^{j}]x^{i} $ as shown in \eqref{risk_ind}.

According to \eqref{risk_ind} and \eqref{rc_s_inner}, the problem of constructing $\Delta$-risk contour \eqref{rc_s}, reduces to the problem of finding an upper bound probability and an upper bound polynomial indicator function of the superlevel set of $p(\mathbf{x},\omega)$. In \cite{Contour}, to compute the upper bound polynomial indicator functions, an $(n_x+n_{\omega})$-dimensional convex optimization problem in the form of a semidefinite program is provided. Such optimization is not suitable  for  online  computations and is limited  to small dimensions $(n_x+n_{\omega})$.

\subsubsection{Analytical Method}
We propose an optimization-free approach to obtain an upper bound on the probability, which is suitable for online computations and large scale problems.
The key insight of this method is to use concentration inequalities to upper bound the risk, obtaining a risk contour involving equations of moments of the representing polynomials of obstacles.  
The computation involved is just the four basic arithmetic operations plus exponentiation. 

Given the polynomial $\mathcal{P}(\mathbf{x},\mathbf{\omega})$ of the uncertain obstacle $\mathcal{X}_{obs}(\omega)$, we define the set $\hat{\mathcal{C}}_{r}^{\Delta}$ as follows:
\begin{align}\label{rc_s_cheby}
\hat{\mathcal{C}}^{\Delta}_{r}= \left\lbrace \ \mathbf{x} \in \mathcal{X}: 
\begin{array}{cc} \frac{\Exp[\mathcal{P}^2(\mathbf{x},\omega)] - \Exp[\mathcal{P}(\mathbf{x},\omega)]^2}{\Exp[\mathcal{P}^2(\mathbf{x},\omega)]} \leq \Delta, \\ 
\Exp[\mathcal{P}(\mathbf{x},\omega)]\leq 0 \end{array}
\right\rbrace
\end{align}
where the expectation is taken with respect to the distribution of uncertain parameter $\omega$. Note that we can compute polynomials $\Exp[\mathcal{P}^2(\mathbf{x},\omega)]$ and $\Exp[\mathcal{P}(\mathbf{x},\omega)]$ in terms of $\mathbf{x}$ and known moments of $\omega$. More precisely, $\Exp[\mathcal{P}^2(\mathbf{x},\omega)]$ and $\Exp[\mathcal{P}(\mathbf{x},\omega)]$ are polynomials in $\mathbf{x}$ whose coefficients are defined in terms of the moments of $\omega$ and the coefficients of polynomial $\mathcal{P}(\mathbf{x},\omega)$.

The following result holds true.\\

\textbf{Theorem 1:} The set $\hat{\mathcal{C}}_{r}^{\Delta}$ in \eqref{rc_s_cheby} is an inner approximation of the static risk contour ${\mathcal{C}}_{r}^{\Delta}$ in \eqref{rc_s}.

\textit{Proof}: 
Let $\mathcal{X}_{obs}(\omega)= \{ \mathbf{x}\in \mathcal{X}: \mathcal{P}(\mathbf{x},\mathbf{\omega}) \geq 0  \}$ be the given static uncertain obstacle as defined in \eqref{obs_s}. To obtain an upper bound of the probability  $\hbox{Prob}( \mathbf{x} \in \mathcal{X}_{obs}(\omega))=\hbox{Prob}( \mathcal{P}(\mathbf{x},\omega)\geq 0 )$, we begin by defining a new random variable $z \in \mathbb{R}$ as follows:
\begin{equation}
z=\mathcal{P}(\mathbf{x},\omega)
\end{equation}
Note that $z \in \mathbb{R}$ is a random variable, while $\mathcal{P}(\mathbf{x},\omega): \mathbb{R}^{n_x+n_{\omega}} \rightarrow \mathbb{R}$ is a polynomial in $\mathbf{x} \in \mathbb{R}^{n_x}$ and random vector $\mathbf{\omega} \in \mathbb{R}^{n_{\omega}}$. By doing so, we can transform the $(n_x+n_{\omega})$-dimensional probability assessment problem into a one-dimensional probability assessment problem in terms of the new defined random variable $z$, i.e., $\hbox{Prob}( \mathbf{x} \in \mathcal{X}_{obs}(\omega))=\hbox{Prob}( \mathcal{P}(\mathbf{x},\omega)\geq 0 )=\hbox{Prob}( z \geq 0 )$, \cite{Risk_Ind}. Note that the statistics of the random variable $z$, e.g., moments, are functions of $\mathbf{x}$ and the statistics of the random vector $\omega$.

Now, to compute an upper bound of the probability $\hbox{Prob}( z \geq 0 )$, we just need an upper bound polynomial description of one-dimensional indicator function $\mathbb{I}_{z\geq0}$ defined as $\mathbb{I}_{z\geq0} = 1 $ if $z\geq 0$, and
0 otherwise. In this paper, we will use the upper bound polynomial indicator function and the upper bound probability provided by
\textit{Cantelli's inequality} defined for scalar random variables as 
\begin{align}
    \hbox{Prob}(z \geq 0) \leq \frac{\Exp[z^2] - \Exp[z]^2}{\Exp[z^2]},  \text{whenever } \Exp[z] \leq 0.
\end{align}
In other words, the probability of collision, i.e., $\hbox{Prob}(z\geq0)=\hbox{Prob}( \mathcal{P}(\mathbf{x},\omega)\geq 0 )$, is bounded if the expected value of remaining safe is nonnegative, i.e, $\Exp[z]=\Exp[\mathcal{P}(\mathbf{x},\omega)] \leq 0$. For other different one-dimensional indicator function-based probability bounds see \cite{Risk_Ind,Contour2,nemirovski2007convex}.

Hence, the upper bound of the probability of collision can be described in terms of the polynomial of the uncertain obstacle as follows:
\begin{align}\label{cheby2}
\hbox{Prob}( \mathbf{x} \in \mathcal{X}_{obs}(\omega)) \leq \frac{\Exp[\mathcal{P}^2(\mathbf{x},\omega)] - \Exp[\mathcal{P}(\mathbf{x},\omega)]^2}{\Exp[\mathcal{P}^2(\mathbf{x},\omega)]}
\end{align}
whenever $\Exp[\mathcal{P}(\mathbf{x},\omega)]\leq 0$. 
This will results in an inner approximation of the $\Delta$-risk contour as in \eqref{rc_s_cheby}. 
Note that although the standard Cantelli's inequality uses the first two moments of the scalar random variable $z$, we need higher order moments of random vector $\omega$ to construct the set in \eqref{rc_s_cheby}. 
In fact, in this paper, we generalize the standard \textit{scalar} Cantelli probability bound to obtain multivariate probability bound \eqref{cheby2} involving nonconvex and nonlinear sets of obstacles.
\hfill $\blacksquare$

\textit{\textbf{Remark 1}}: The set in \eqref{rc_s_cheby} is a \textit{rational} polynomial-based inner approximation of the risk contour in \eqref{rc_s}. It also uses higher order moments of the uncertain parameter $\omega$ up to order $2d$, where $d$ is the order of the polynomial obstacle $\mathcal{P}(\mathbf{x},\omega)$. 
Since $\hat{\mathcal{C}}^{\Delta}_{r}$ is an inner approximation of ${\mathcal{C}}^{\Delta}_{r}$, any trajectory $\mathbf{x}(t) \in \hat{\mathcal{C}}^{\Delta}_{r}, \forall t\in [t_0,t_f]$, is guaranteed to have a risk less or equal to $\Delta$.
\hfill \qedsymbol

We now provide an illustrative example to show the performance of the proposed method to construct the static $\Delta$-risk contours and benchmark our method against the optimization-based approach in \cite{Contour}.

\begin{figure}
    \centering
    \includegraphics[scale=0.22]{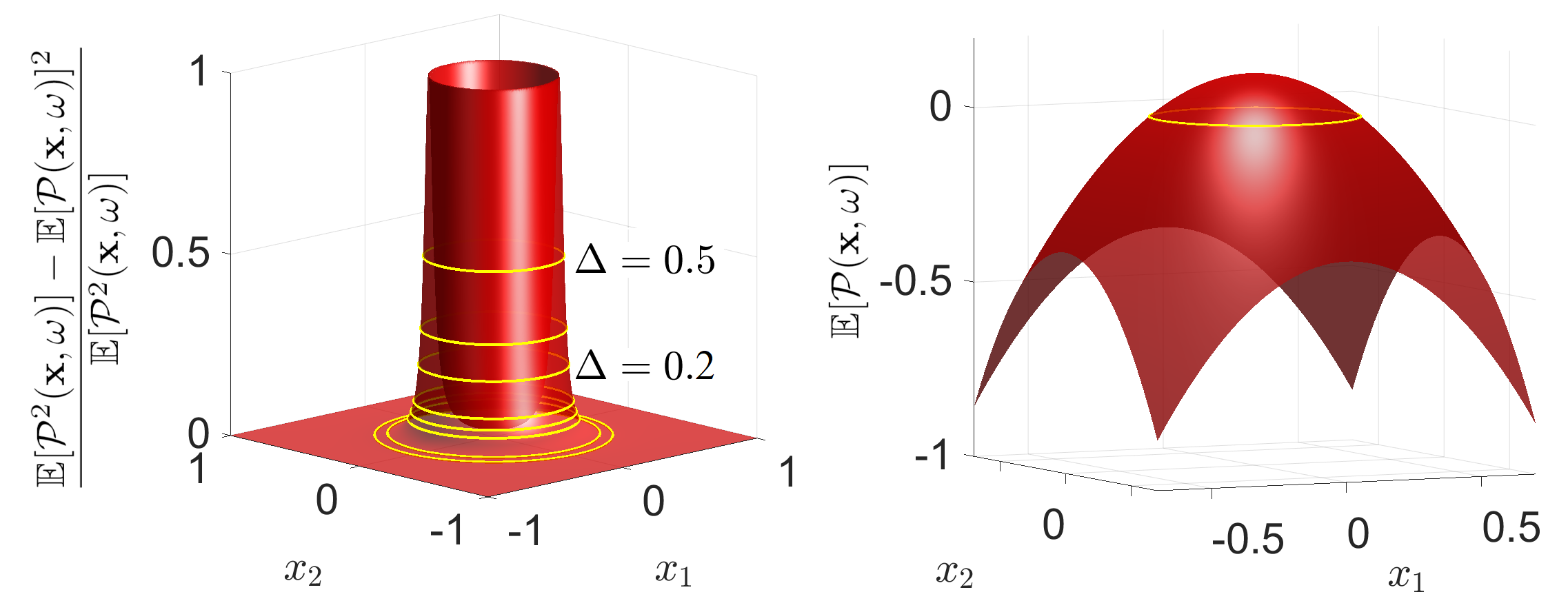}
	\caption{{Illustrative example 1: Intersection of $\Delta$-sublevel set of function $\frac{\Exp[\mathcal{P}^2(\mathbf{x},\omega)] - \Exp[\mathcal{P}(\mathbf{x},\omega)]^2}{\Exp[\mathcal{P}^2(\mathbf{x},\omega)]}$ and $0$-sublevel set of function $\Exp[\mathcal{P}(\mathbf{x},\omega)]$ describes the static $\Delta$-risk contour as in \eqref{rc_s_cheby}.}}
    \label{fig_1}
\end{figure}

\begin{figure}
    \centering
  \includegraphics[scale=0.24]{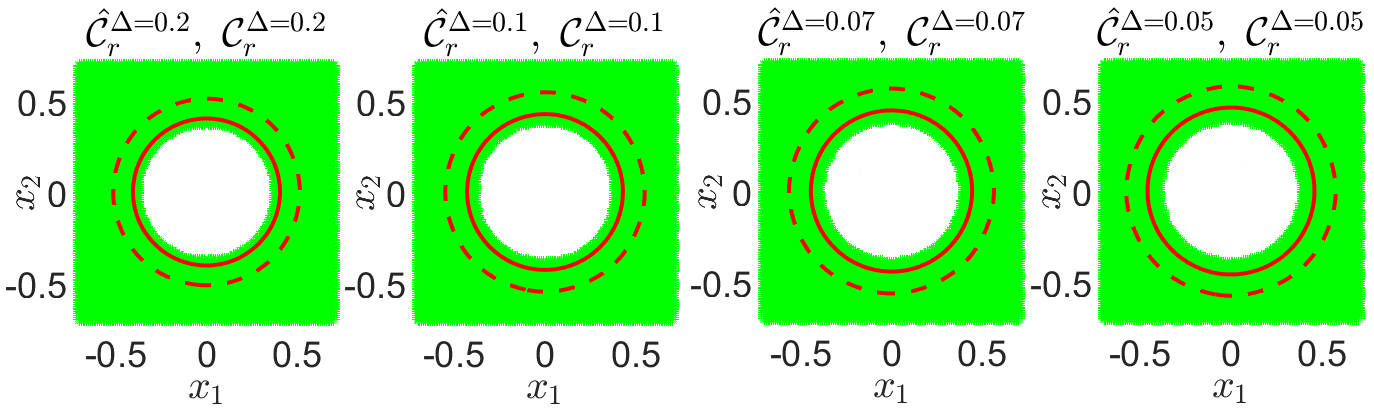}   
	\caption{{Illustrative example 1: True static $\Delta$-risk contour ${\mathcal{C}}^{\Delta}_{r}$ (green) and inner approximation $\hat{\mathcal{C}}^{\Delta}_{r}$ obtained using i) the proposed analytical method in \eqref{rc_s_cheby} (outside of the solid-line) and ii) the proposed optimization-based method in (\cite{Contour}, Fig.4 ) (outside of the dashed-line). While the optimization-based method in \cite{Contour} uses a standard polynomial of order 20, the proposed analytical method in \eqref{rc_s_cheby} uses 4th order rational and 2nd order standard polynomials to describe the risk contours.}}
    \label{fig_2}
\end{figure}

\textbf{Illustrative Example 1:
}
Consider the following illustrative example where $\mathcal{X}=[-1,1]^2$. The set {$\mathcal{X}_{obs}(\omega)= \left\lbrace  (x_1,x_2) :  \omega^2-x_1^2-x_2^2 \geq 0 \right\rbrace $} represents a circle-shaped obstacle 
whose radius $\omega$ has a uniform probability distribution over {$[0.3,0.4]$}, \cite{Contour}. 
Moment of order $\alpha$ of a uniform distribution defined over $[l,u]$ can be described in a closed-form as $\frac{u^{\alpha+1}-l^{\alpha+1}}{(u-l)(\alpha+1)}$.
To construct the static $\Delta$-risk contour, we compute polynomials $\Exp[\mathcal{P}(\mathbf{x},\omega)]$
and $\Exp[\mathcal{P}^2(\mathbf{x},\omega)]$ using the polynomial obstacle and the moments of $\omega$ as follows:
\begin{align}
    \Exp[\mathcal{P}(\mathbf{x},\omega)] &= \Exp[\omega^2-x_1^2-x_2^2 ] \notag\\
    &=\Exp[\omega^2]-x_1^2-x_2^2 =0.1233-x_1^2-x_2^2 \notag\\
     \Exp[\mathcal{P}^2(\mathbf{x},\omega)] &= \Exp[\left(\omega^2-x_1^2-x_2^2 \right)^2]  \notag \\ 
    &=\Exp[\omega^4] - 2\Exp[\omega^2]x_1^2 - 2\Exp[\omega^2]x_2^2 \notag\\
    & \ \ \ \ + x_1^4 + 2x_1^2x_2^2 + x_2^4\notag \\
    &=0.0156 - 0.2466x_1^2 - 0.2466x_2^2 \notag\\
    & \ \ \ \ + x_1^4 + 2x_1^2x_2^2 + x_2^4 \notag
\end{align}
As shown in Figures \ref{fig_1} and \ref{fig_2}, we use the sublevels of functions $\frac{\Exp[\mathcal{P}^2(\mathbf{x},\omega)] - \Exp[\mathcal{P}(\mathbf{x},\omega)]^2}{\Exp[\mathcal{P}^2(\mathbf{x},\omega)]} $ and $\Exp[\mathcal{P}(\mathbf{x},\omega)]$ as in \eqref{rc_s_cheby} to construct the inner approximations of the static $\Delta$-risk contours for different risk levels $\Delta=[0.2,0.1,0.07,0.05]$.
We also compare our proposed method in  \eqref{rc_s_cheby} with the optimization-based method in \cite{Contour} as shown in Figure \ref{fig_2}. 
Using the provided SDP in \cite{Contour}, we obtain a polynomial of order 20 to describe the inner approximations of the $\Delta$-risk contours. We note that the proposed analytical method of this paper provides a tight inner approximation of the $\Delta$-risk contours. This is primarily due to the facts that i) the proposed analytical approach results in a \textit{rational} polynomial representation of the risk-contours as opposed to a \textit{standard} polynomial representation provided in \cite{Contour} and ii) with the provided analytical approach, we are able to avoid the numerical issues that arise when solving large-scale SDPs \cite{roux2018validating}. The proposed method of this paper is also suitable for online large scale planning problems. 

\subsubsection{Comparison with Linear Gaussian Method}\label{chap4_comparison_lingaussian_1}
We compare our method with the linear Gaussian method proposed in \cite{blackmore2011chance}. 
In their setting, the system dynamics is discrete-time linear time invariant system with Gaussian noise. 
The obstacles are represented by convex polytopes. 
In this chapter, we compare our polynomial representation of risk contours to the convex polytopic representation of the obstacle in the linear Gaussian method. 

To use the linear Gaussian method, we need to find, for each obstacle, a risk contour represented as a union of half planes
\begin{align}\label{obstacle_lin_gaussian}
    \widetilde{\mathcal{C}}_r^{\Delta} = \bigcup_{i=1}^\ell \{\tbx \in \mathcal{X}: \mathbf{a}_i^\top \tbx \leq b_i \}.
\end{align}
In principle, we can obtain $\widetilde{\mathcal{C}}_r^{\Delta}$ either from the original obstacle representation ${\mathcal{C}}_r^{\Delta}$ such that $\widetilde{\mathcal{C}}_r^{\Delta} \subseteq {\mathcal{C}}_r^{\Delta}$, or from the polynomial risk contour $\hat{\mathcal{C}}_r^{\Delta}$ obtained from the analytical method such that $\widetilde{\mathcal{C}}_r^{\Delta} \subseteq \hat{\mathcal{C}}_r^{\Delta} \subseteq {\mathcal{C}}_r^{\Delta}$.
Given a polynomial representation of ${\mathcal{C}}_r^{\Delta}$ involving random variables as in \eqref{rc_s}, it is in general impossible to directly compute the half planes $\mathbf{a}_i^\top \tbx \leq b_i$ of $\widetilde{\mathcal{C}}_r^{\Delta}$.
Nevertheless, in rare cases where the representation of ${\mathcal{C}}_r^{\Delta}$ is very simple, it is possible to obtain $\widetilde{\mathcal{C}}_r^{\Delta}$ directly from ${\mathcal{C}}_r^{\Delta}$ as shown in the simple example below.

\begin{figure}[t]
    \centering
    \includegraphics[width=0.2\textwidth]{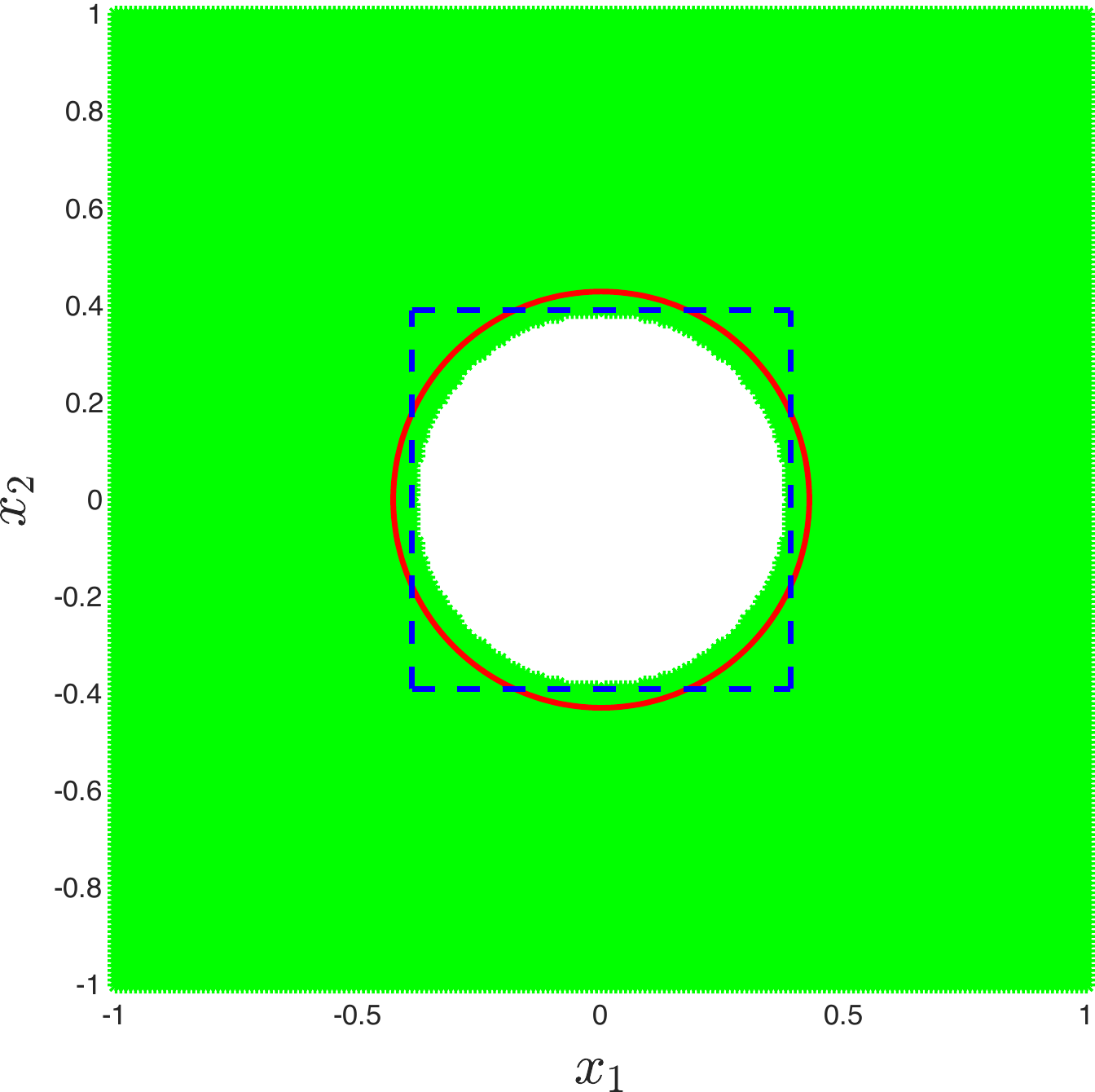}
    \caption{A simple example where $\widetilde{\mathcal{C}}_r^{\Delta}$ can be directly calculated from ${\mathcal{C}}_r^{\Delta}$. The green region is the true $\Delta$-risk contour ${\mathcal{C}}_r^{\Delta}$, obtained from Monte Carlo simulation. The red circle is boundary of the $\Delta$-risk contour $\hat{\mathcal{C}}_r^{\Delta}$ obtained by our analytical method. The blue dashed square is the boundary of the $\Delta$-risk contour of $\widetilde{\mathcal{C}}_r^{\Delta}$.}
    \label{fig:thesis_chp4_cmp_exp1}
\end{figure}

\textbf{A Simple Example.} Consider Illustrative Example 1.
The state space is $\mathcal{X}=[-1,1]^2$. 
The set $\mathcal{O}(\omega)= \{  (x_1,x_2) \in \mathcal{X} :  \omega^2-x_1^2-x_2^2 \geq 0 \}$ represents a circle-shaped obstacle whose radius $\omega$ has a uniform probability distribution over $[0.3,0.4]$.
Suppose we want to use a rectangle whose sides are parallel to the axis as the boundary of $\widetilde{\mathcal{C}}_r^{\Delta}$.
The square $S(\omega) = \{(x_1, x_2) \in \mathcal{X}: -\omega \leq x_1, x_2 \leq \omega\}$ is an outer approximation of the obstacle $\mathcal{O}(\omega)$ for any $\omega \in [0.3,0.4]$.
So $\mathcal{X} \backslash S(\omega) =  \{(x_1,x_2) \in \mathcal{X} : x_1 < -\omega \} \cup \{(x_1,x_2) \in \mathcal{X} :  x_2 < -\omega \} \cup \{ (x_1,x_2) \in \mathcal{X} : x_1 > \omega\} \cup \{(x_1,x_2) \in \mathcal{X} : x_2 > \omega\}$.
Suppose $\Delta = 0.1$, then $\widetilde{\mathcal{C}}_r^{\Delta}$ can be calculated from $\mathcal{X} \backslash S(\omega)$ as $\{(x_1,x_2) \in \mathcal{X} : x_1 < -0.39 \} \cup \{(x_1,x_2) \in \mathcal{X} :  x_2 < -0.39 \} \cup \{ (x_1,x_2) \in \mathcal{X} : x_1 > 0.39\} \cup \{(x_1,x_2) \in \mathcal{X} : x_2 > 0.39\}$. 
The boundary of $\widetilde{\mathcal{C}}_r^{\Delta}$ is plotted as a dashed square in \Cref{fig:thesis_chp4_cmp_exp1}.

\begin{figure}[t]
    \centering
    \includegraphics[width=0.23\textwidth]{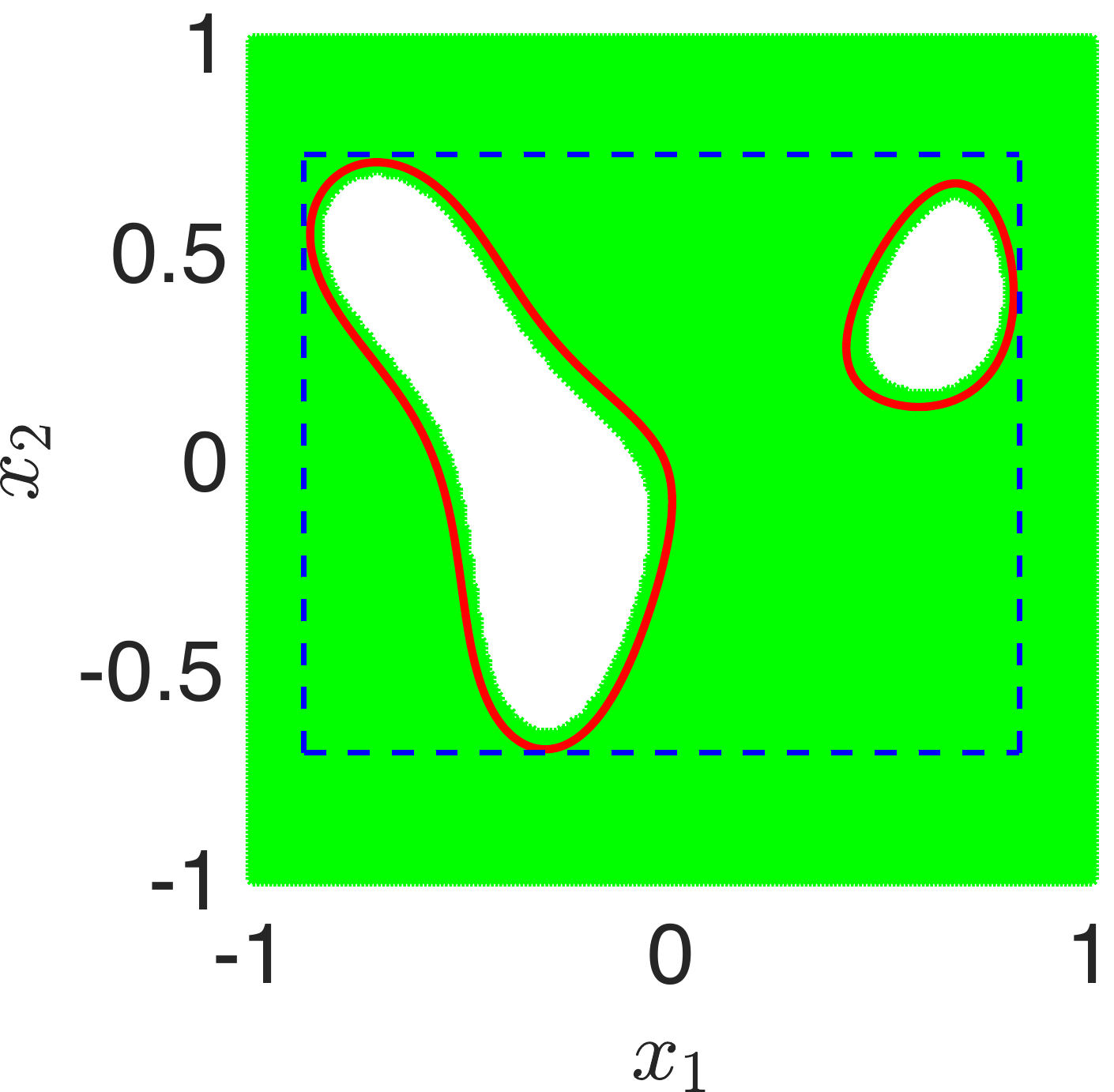}
    \includegraphics[width=0.23\textwidth]{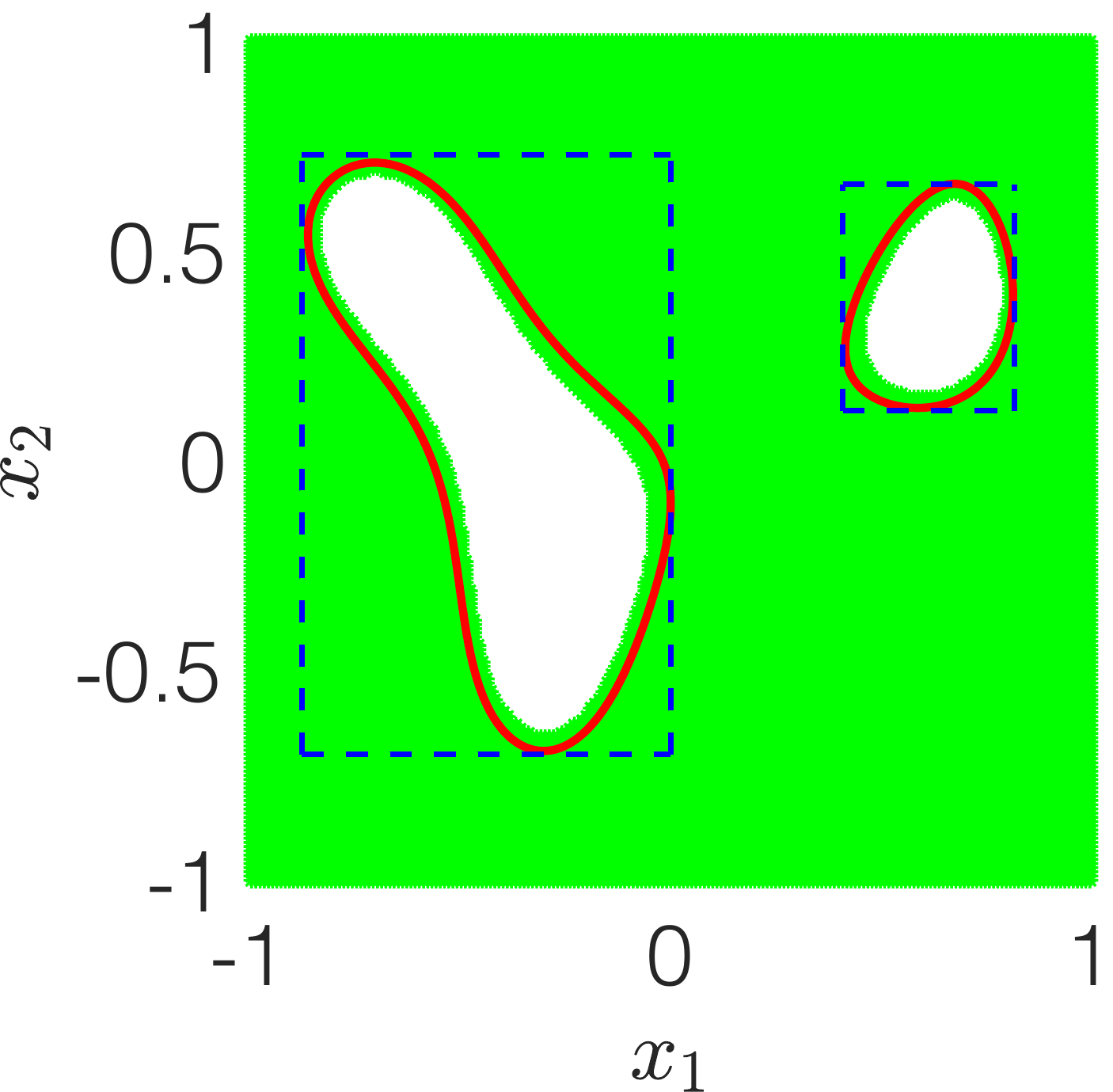}
    \caption{A more complicated example where $\widetilde{\mathcal{C}}_r^{\Delta}$ is obtained from $\hat{\mathcal{C}}_r^{\Delta}$. The green region is the true $\Delta$-risk contour ${\mathcal{C}}_r^{\Delta}$, obtained from Monte Carlo simulation. The red curves are the boundary of the $\Delta$-risk contour $\hat{\mathcal{C}}_r^{\Delta}$ obtained by our analytical method. The blue dashed rectangle is the boundary of the $\Delta$-risk contour of $\widetilde{\mathcal{C}}_r^{\Delta}$. 
    Left: One-rectangle approximation.
    Right: Two-rectangle approximation.
    }
    \label{fig:thesis_chp4_cmp_exp2}
\end{figure}

Apart from the simple example above, $\widetilde{\mathcal{C}}_r^{\Delta}$ can rarely be directly calculated from ${\mathcal{C}}_r^{\Delta}$.
In those non-trivial cases, we have to obtain $\hat{\mathcal{C}}_r^{\Delta}$ first and then inner approximate it to get $\widetilde{\mathcal{C}}_r^{\Delta}$. 
The resulting approximation can be very conservative as shown in the example below.

\textbf{A More Complicated Example.} Consider Example \ref{sec:exp_a1}, where the uncertain obstacle is represented by a 5th-order polynomial and the noise has Beta distribution.
Suppose we want to use a rectangle whose sides are parallel to the axis as the boundary of $\widetilde{\mathcal{C}}_r^{\Delta}$.
As shown in \Cref{fig:thesis_chp4_cmp_exp2} Left, the boundary of $\widetilde{\mathcal{C}}_r^{\Delta}$ is plotted in dashed lines and it is so conservative that it excludes the feasible region between two components of the obstacle.
In order to be less conservative, we use two rectangles to approximate the feasible region as shown in \Cref{fig:thesis_chp4_cmp_exp2} Right.

In summary, given a polynomial representation of an obstacle, the risk contour of the linear Gaussian method is in general more conservative than that of our method. 
Since in non-trivial cases, we have to obtain $\hat{\mathcal{C}}_r^{\Delta}$ first and then inner approximate it to get $\widetilde{\mathcal{C}}_r^{\Delta}$, the linear Gaussian method always takes more computational time.
In very simple and rare cases, the risk contour of the linear Gaussian method can be calculated directly from the obstacle representation and is not too conservative. 

\subsection{Dynamic Risk Contours} 
Let $\mathcal{X}_{obs}(\omega,t)= \{ \mathbf{x}\in \mathcal{X}: \mathcal{P}(\mathbf{x},\mathbf{\omega},t) \geq 0  \}$ be the given dynamic uncertain obstacle as defined in \eqref{obs_d} and $\Delta \in [0,1]$ be the given acceptable risk level in \eqref{opt_prob2}. Then, dynamic $\Delta$-risk contour at time $t$ denoted by $\mathcal{C}^{\Delta}_{r}(t)$ is defined as the set of all points in the environment, i.e., $\mathbf{x} \in \mathcal{X}$, whose risk at time $t$ is less or equal to $\Delta$. More precisely,
\begin{equation}\label{rc_d}
\mathcal{C}^{\Delta}_{r}(t):= \{\ \mathbf{x} \in \mathcal{X}: \ \hbox{Prob}( \mathbf{x} \in \mathcal{X}_{obs}(\omega,t)) \leq \Delta  \}
\end{equation}

Similar to the static risk contours, we can replace the probabilistic constraint in \eqref{rc_d} with deterministic constraints and construct an inner approximation of the dynamic $\Delta$-risk contour denoted by $\hat{\mathcal{C}}^{\Delta}_{r}(t)$ as follows:
\begin{align}\label{rc_d_cheby}
	\hat{\mathcal{C}}^{\Delta}_{r}(t)= \left\lbrace \ \mathbf{x} \in \mathcal{X}: 
\begin{array}{cc} \frac{\Exp[\mathcal{P}^2(\mathbf{x},\omega,t)] - \Exp[\mathcal{P}(\mathbf{x},\omega,t)]^2}{\Exp[\mathcal{P}^2(\mathbf{x},\omega,t)]} \leq \Delta, \\ 
\Exp[\mathcal{P}(\mathbf{x},\omega,t)]\leq 0  \end{array}
\right\rbrace
\end{align}
Note that dynamic $\Delta$-risk contour \eqref{rc_d_cheby} is described in terms of the time-varying constraints.

\begin{figure}
    \centering
    \includegraphics[scale=0.25]{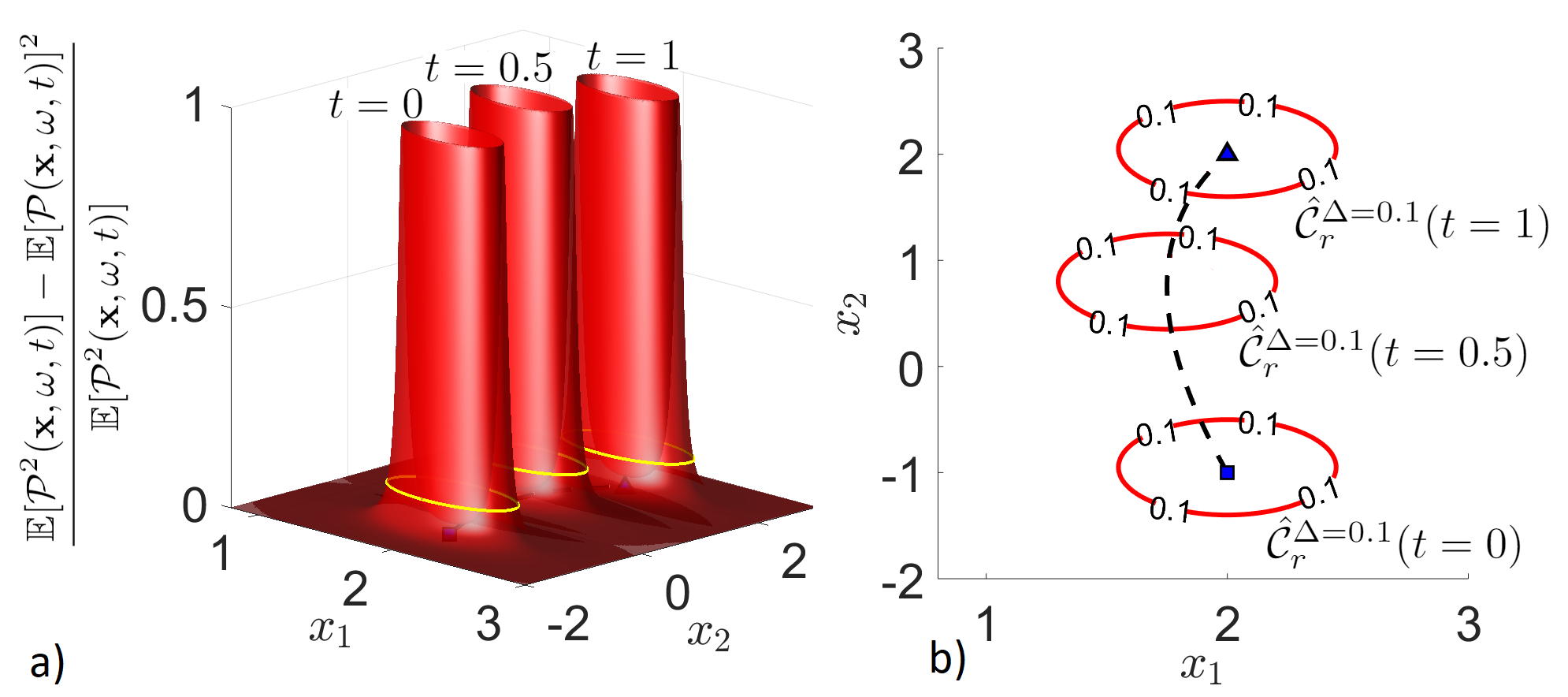}
	\caption{{Illustrative example 2:
	a) Function $\frac{\Exp[\mathcal{P}^2(\mathbf{x},\omega,t)] - \Exp[\mathcal{P}(\mathbf{x},\omega,t)]^2}{\Exp[\mathcal{P}^2(\mathbf{x},\omega,t)]}$ at time steps $t=0,0.5,1$, b) Dynamic $\Delta$-risk contours for $\Delta=0.1$ at time steps $t=0,0.5,1$ 
	described in \eqref{rc_d_cheby}. Dashed line shows the expected value of the given uncertain trajectory, i.e., $\Exp[(p_{x_1}(t,\omega_2),p_{x_2}(t,\omega_3))]$. At each time $t$, for any point inside $\hat{\mathcal{C}}^{\Delta}_{r}(t)$ (outside of the closed curve), probability of collision with the moving uncertain obstacle is less or equal to $\Delta=0.1$. }}
    \label{fig_3}
\end{figure}

\textbf{Illustrative Example 2:}
Consider the following illustrative example where $\mathcal{X}=[-1,1]^2$. The set {$\mathcal{X}_{obs}(\omega,t)= \{  (x_1,x_2) :  \omega_1^2-(x_1-p_{x_1}(t,\omega_2))^2-(x_2-p_{x_2}(t,\omega_3))^2 \geq 0 \}$}  represents a moving circle-shaped obstacle with uncertain radius $\omega_1$ and uncertain trajectories $ p_{x_1}(t,\omega_2)=2-t+t^2+0.2\omega_2,\  p_{x_2}(t,\omega_3)=-1+4t-t^2+0.1\omega_3 $ that describe the uncertain motion of the obstacle over the time horizon $t\in[0,1]$. Uncertain parameters have uniform, normal, and Beta distributions as $\omega_1 \sim Uniform[0.3,0.4]$, $\omega_2 \sim \mathcal{N}(0,0.1)$, $\omega_3 \sim Beta(3,3)$. Moment of order $\alpha$ of a Beta distribution with parameters $(a,b)$ 
and a normal distribution with mean $\mu$ and standard deviation $\sigma$ can be described in closed-forms as {$y_{\alpha}=\frac{a+\alpha-1}{a+b+\alpha-1}y_{\alpha-1}, y_0=1$} and {$y_{\alpha}=\sigma^{\alpha}(-\sqrt{-1}\sqrt{2})^{\alpha}kummerU(\frac{-\alpha}{2}, \frac{1}{2}, \frac{-\mu^2}{2\sigma^2})$}, respectively, where { $kummerU$} is Kummer's (confluent hyper-geometric) function \cite{kummer1837integralibus}. 
Similar to illustrative example 1, we compute the polynomials $\Exp[\mathcal{P}^2(\mathbf{x},\omega,t)]$ and 
$\Exp[\mathcal{P}(\mathbf{x},\omega,t)]$ using the moments of the uncertain parameters $\omega_i, i=1,2,3$ and the polynomial obstacle. We then construct the dynamic $\Delta$-risk contour as a function of time as described in \eqref{rc_d_cheby}.
Figure \ref{fig_3} shows the obtained dynamic $\Delta$-risk contours for $\Delta=0.1$ at time steps $t=0,0.5,1$ along the given uncertain trajectory $\left( p_{x_1}(t,\omega_2),\  p_{x_2}(t,\omega_3) \right)$.

\textit{\textbf{Remark 2}}: We can use \eqref{rc_s_cheby}
and \eqref{rc_d_cheby} to construct static and dynamic risk contours in real-time.
Therefore, standard motion planning algorithms such as RRT$^*$, PRM, and motion primitive-based methods can be used for real-time risk bounded motion planning. To accomplish this, one just needs
to use the safe regions, i.e, risk contours, to construct the trajectories.  \hfill \qedsymbol

In the next section, we provide \textit{continuous-time} planning algorithms to look for trajectories with guaranteed bounded risk over the entire planning time horizon without the need for \textit{time discretization}.

\section{Continuous-time risk bounded trajectory planning using risk contours} \label{sec_plan}
In this section, we will use the static and dynamic risk contours to solve the continuous-time risk bounded trajectory planning problem defined in \eqref{opt_obj}. More precisely, we use the static and dynamic risk contours to transform the probabilistic optimization in \eqref{opt_obj} into a deterministic polynomial optimization. 
The obtained deterministic polynomial optimization is, in general, nonconvex and nonlinear. 
In addition, we need to ensure that all the obtained deterministic constraints are satisfied over the entire planning time horizon $[t_0,t_f]$. In this section, we provide convex methods based on SOS techniques introduced in Section \ref{sec_def} to efficiently solve the obtained nonconvex time-varying deterministic planning optimization problem. While the existing planners rely on time discretization to verify the planning safety constraints, the provided SOS-based planners look for continuous-time trajectories with guaranteed bounded risk over the entire planning time horizon without the need for time discretization.

We first begin by addressing the continuous-time risk bounded trajectory planning in static uncertain environments using the static risk contours. We then use the dynamic risk contours to address the planning problems in dynamic uncertain environments.

\subsection{Planning in Static Uncertain Environments}

In this section, we are concerned with continuous-time risk bounded trajectory planning in the presence of static uncertain obstacles of the form \eqref{obs_s}. More precisely, we aim at solving 
the probabilistic optimization problem in \eqref{opt_obj} in the presence of static uncertain obstacles $\mathcal{X}_{obs_i}(\omega_i), i=1,...,n_{o_s}$, i.e.,

\begin{align} 
& \underset{\mathbf{x}(t): [t_0, t_f] \rightarrow \mathbb{R}^{n_x}}{\text{minimize}}
 \int_{t_0}^{t_f}  \left\Vert \dot{\mathbf{x}}(t) \right\Vert_2^2 dt \label{opt_s_obj} \\
& \text{subject to} \ 
\ \ \mathbf{x}(t_0)=\mathbf{x}_0, \  \mathbf{x}(t_f)=\mathbf{x}_f \tag{\ref{opt_s_obj}a} \\
&  \ \hbox{Prob}\left( \mathbf{x}(t) \in \mathcal{X}_{obs_i}(\omega_i)    \right) \leq \Delta, \ \forall t\in [t_0,t_f], \  i=1,...,n_{o_s} \tag{\ref{opt_s_obj}b} \label{opt_s_prob1} 
\end{align}
To obtain the deterministic polynomial optimization, we replace probabilistic constraints \eqref{opt_s_prob1} with deterministic constraints in terms of the static $\Delta$-risk contours as follows:

\begin{align} 
& \underset{\mathbf{x}(t): [t_0, t_f] \rightarrow \mathbb{R}^{n_x}}{\text{minimize}}
 \int_{t_0}^{t_f}  \left\Vert \dot{\mathbf{x}}(t) \right\Vert_2^2 dt \label{opt_s2_obj} \\
& \text{subject to} \ 
\ \ \mathbf{x}(t_0)=\mathbf{x}_0, \  \mathbf{x}(t_f)=\mathbf{x}_f \tag{\ref{opt_s2_obj}a} \\
&  \mathbf{x}(t) \in \hat{\mathcal{C}}^{\Delta}_{r_i}   , \ \forall t\in [t_0,t_f], \  i=1,...,n_{o_s} \tag{\ref{opt_s2_obj}b} \label{opt_s2_prob1} 
\end{align}
where $\hat{\mathcal{C}}^{\Delta}_{r_i}$ is the static $\Delta$-risk contour of the static uncertain obstacle $\mathcal{X}_{obs_i}(\omega_i)$. The set of $\hat{\mathcal{C}}^{\Delta}_{r_i}, i=1,...,n_{o_s}$ represents the inner approximation of the feasible set of the probabilistic optimization in \eqref{opt_s_obj}. The obtained deterministic optimization in \eqref{opt_s2_obj} is time-varying optimization problem where we need to ensure to satisfy the constraints over the entire planning time horizon $[t_0,t_f]$. 

To solve the deterministic optimization in \eqref{opt_s2_obj}, we will look for i) polynomial trajectory defined in \eqref{poly} and ii) piece-wise linear trajectory defined in \eqref{linear}. 
By substituting the polynomial trajectory $\mathbf{x}(t)=\sum_{\alpha=0}^d \mathbf{c}_{\alpha}t^{\alpha}$ in optimization \eqref{opt_s2_obj}, we obtain a deterministic optimization with constraints in terms of the coefficients of the polynomial trajectory $\mathbf{x}(t)$, i.e., 
	$(1-\Delta)\Exp[\mathcal{P}^2(\sum_{\alpha=0}^d \mathbf{c}_{\alpha}t^{\alpha},\omega)] - \Exp[\mathcal{P}(\sum_{\alpha=0}^d \mathbf{c}_{\alpha}t^{\alpha},\omega)]^2 \leq 0$, $ \Exp[\mathcal{P}(\sum_{\alpha=0}^d \mathbf{c}_{\alpha}t^{\alpha},\omega)]\leq 0$.
Similarly, by substituting the piece-wise linear trajectory of \eqref{linear}, we obtain the objective function $\sum_{j=1}^s \int_{t_{j-1}}^{t_{j}}  \left\Vert \dot{\mathbf{x}}_j(t) \right\Vert_2^2 dt $
and constraints 
$ \mathbf{a}_j+ \mathbf{b}_jt \in \hat{\mathcal{C}}^{\Delta}_{r_i}, \ \forall t\in [t_{j-1},t_j], \ i=1,...,n_{o_s}, j=1,...,s $. To solve the obtained time-varying deterministic polynomial optimization problems, we will provide 3 methods based on SOS optimization techniques as follows:

\subsubsection{Time-Varying SOS Optimization}\label{sec_s_plan}
we use time-varying SOS optimization, introduced in Section \ref{sec_def}, to solve the time-varying deterministic polynomial optimization in \eqref{opt_s2_obj}.
In implementation, we use the heuristic algorithm introduced by \cite{SOS_time2}.

\subsubsection{Standard SOS Optimization}
In this method, we obtain a standard polynomial optimization of the form \eqref{nlp} by eliminating time $t$. We then use the standard SOS optimization technique in \eqref{nlp_sos} to solve the obtained optimization problem. To achieve this, let $\mathcal{X}_{obs}(\omega)= \{ \mathbf{x}\in \mathcal{X}: \mathcal{P}(\mathbf{x},\mathbf{\omega}) \geq 0  \}$ be the given static uncertain obstacle defined in \eqref{obs_s} and $\mathbf{x}(t)$ be the polynomial trajectory in \eqref{poly}. To eliminate time $t$, instead of using the instant risk as in \eqref{opt_s_prob1}, we work with the average risk defined as 
$\frac{1}{t_f-t_0} \int \int_{ \{ (t,\mathbf{\omega}): \mathcal{P}(\mathbf{x}(t),\omega)\geq 0\}} pr(\mathbf{\omega})d\mathbf{\omega}dt$ where $pr(\mathbf{\omega})$ is the probability density function of $\omega$. This is equivalent to treating $t$ as a random variable with a uniform probability distribution over the planning time horizon $[t_0,t_f]$.
We should note that the average risk is a \textit{weak} safety measure, which means that just by bounding the average risk, we cannot guarantee to satisfy the constraints of the probabilistic optimization in \eqref{opt_s_obj}.

By defining the average risk, we substitute the trajectory $\mathbf{x}(t)$ in the probabilistic constraint and follow the same steps as in Section \ref{sec_rc_s} to obtain an upper bound of the average risk and construct the set of all coefficients $\mathbf{c}_{\alpha}|_{\alpha =0}^d$ that results in a risk bounded trajectory of the form $\mathbf{x}(t)=\sum_{\alpha=0}^d \mathbf{c}_{\alpha}t^{\alpha}$.
More precisely, we obtain the following set for the coefficients: 
\begin{center} \resizebox{0.99\linewidth}{!}{$
\left\lbrace \ \mathbf{c}_{\alpha}|_{\alpha =0}^d:\begin{array}{cc}(1-\Delta)\Exp[\mathcal{P}^2(\sum_{\alpha=0}^d \mathbf{c}_{\alpha}t^{\alpha},\omega)] - \Exp[\mathcal{P}(\sum_{\alpha=0}^d \mathbf{c}_{\alpha}t^{\alpha},\omega)]^2 \leq 0, \\ \Exp[\mathcal{P}(\sum_{\alpha=0}^d \mathbf{c}_{\alpha}t^{\alpha},\omega)]\leq 0\end{array}
 \right\rbrace
$}
\end{center}
where the expectation is taken with respect to the probability distributions of $\omega$ and $t$. We can obtain a similar set for the coefficients of the piece-wise linear trajectories, as well. By computing such sets for coefficients of the trajectories, we can transform the probabilistic optimization problem into a deterministic standard polynomial optimization and use the standard SOS optimization technique in \eqref{nlp_sos} to obtain the optimal solution. \\

\subsubsection{RRT-SOS}\label{sec_RRT}
In this method, we use sampling-based motion planning algorithms like RRT to construct the risk bounded trajectory of the deterministic polynomial optimization in \eqref{opt_s2_obj}. To ensure safety along the edges of the RRT, we use an SOS-based continuous-time technique to verify the constraints of the optimization in \eqref{opt_s2_prob1} as follows:

Let $\mathbf{x}(t)=\sum_{\alpha=0}^d \mathbf{c}_{\alpha}t^{\alpha}, t\in[t_1,t_2]$ be the given trajectory between the two samples $\mathbf{x}_1 \in \mathcal{X}$ and $\mathbf{x}_2 \in \mathcal{X}$ in the uncertain environment. Also, let $\mathcal{S}=\{\mathbf{x}: g_i(x)\geq 0, i=1,...,\ell \}$ be the feasible set of optimization \eqref{opt_s2_obj}, i.e., the set constructed by the polynomial constraints of all the risk contours $\hat{\mathcal{C}}^{\Delta}_{r_i}, i=1,...,n_{o_s}$. Then, the following result holds true:

Polynomial $\mathbf{x}(t)$ satisfies the safety constraints of the deterministic optimization in \eqref{opt_s2_obj} over the time interval $[t_1,t_2]$, i.e., $\mathbf{x}(t) \in \mathcal{S}$ for all $t\in[t_1,t_2]$, if and only if polynomials $g_i(\mathbf{x}(t)), i=1,...,\ell$ take the following SOS representation:
\begin{align}\label{SOS_Cond}
    g_i(\mathbf{x}(t))={\sigma_0}_i(t)+ {\sigma_1}_i(t)(t-t_1) + {\sigma_2}_i(t)(t_2-t) \ |_{i=1}^{\ell} 
\end{align}
where ${\sigma_0}_i(t), {\sigma_1}_i(t), {\sigma_2}_i(t), i=1,...,\ell$ are SOS polynomials with appropriate degrees \cite{putinar1993positive,SOS2,SOS3}. Yalmip \cite{Yalmip_1} and Spotless \cite{Spot_1} packages can be used to check the SOS condition \eqref{SOS_Cond} for the given trajectory $\mathbf{x}(t)$. 

\textit{\textbf{Remark 3}}: The complexity of the safety SOS condition in \eqref{SOS_Cond} is independent of the size of the planning time horizon $[t_1,t_2]$ and the length of the polynomial trajectory $\mathbf{x}(t)$. Hence, one can use \eqref{SOS_Cond} to easily verify the safety of trajectories in uncertain environments over the long planning time horizons. \hfill \qedsymbol

\begin{figure}
    \centering
    \includegraphics[scale=0.28]{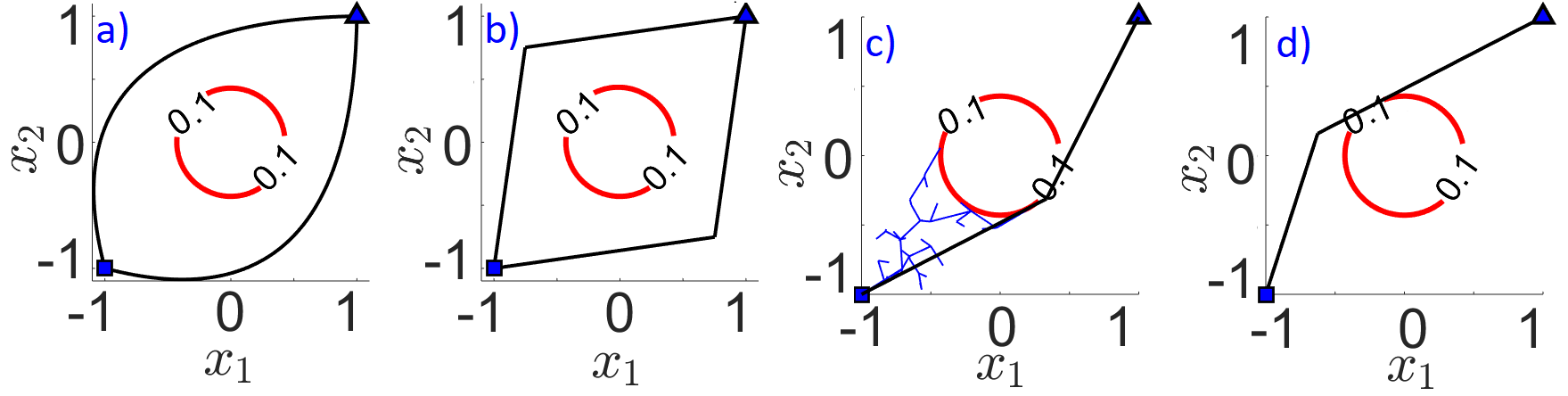}
	\caption{{Illustrative Example 3: a) risk bounded polynomial trajectories of order 2 obtained via standard SOS optimization, b) risk bounded piece-wise linear trajectories obtained via standard SOS optimization, c) risk bounded piece-wise linear trajectory obtained via RRT-SOS algorithm, d) risk bounded piece-wise linear trajectory obtained via time-varying SOS optimization.}}
    \label{fig_illus3_1}
\end{figure}

The safety condition in \eqref{SOS_Cond} can be used in any sampling-based motion planning algorithms to verify the safety of the trajectories between the sample points. In this paper, we will use the following naive RRT algorithm:
To expand the RRT, we use a linear trajectory to connect the given sample point to the closest vertex in the tree if the linear trajectory satisfies the SOS condition in \eqref{SOS_Cond}, i.e., this implies that the linear trajectory between the two points is inside the risk contours in \eqref{opt_s2_prob1}. 
As a termination condition, we also check the safety of the linear trajectory 
between the selected safe sample and the goal points via the SOS condition in \eqref{SOS_Cond}. If the linear trajectory satisfies SOS condition \eqref{SOS_Cond}, we connect the sample point to the goal point; Hence, a feasible trajectory between the start and goal points can be constructed.

To improve the obtained feasible trajectory, we perform the following steps: i) given the obtained tree, we construct a graph, e.g., PRM, whose nodes are the vertex of the tree and edges of the graph are all the linear trajectories between the nodes that satisfy the SOS condition in \eqref{SOS_Cond}, ii) we then perform a shortest path algorithm, e.g.,  Dijkstra algorithm, to obtain a path from the start to the goal point. 
We can also use smart initialization to guide the RRT-SOS algorithm and improve the run-time. For example, we use the straight line between the start and goal points to initialize the  RRT-SOS algorithm. We then perform the sampling in the neighborhood of the given initial path and  incrementally increase the size of the neighborhood until a feasible trajectory is obtained.

\textbf{Illustrative Example 3:} Consider the uncertain obstacle in illustrative example 1. We want to find a risk bounded trajectory between the points $\mathbf{x}(0)=[-1,-1]$ and $\mathbf{x}(1)=[1,1]$ by solving the probabilistic optimization in \eqref{opt_s_obj} with $\Delta=0.1$. For this purpose, we solve the deterministic optimization problem in \eqref{opt_s2_obj} with respect to the $0.1$-risk contour of the uncertain obstacle using the discussed 3 methods as shown in Figure \ref{fig_illus3_1}. More precisely, we use standard SOS optimization to obtain a polynomial trajectory of order 2 and also a piece-wise linear trajectory with 2 pieces. Using GloptiPoly package, we extract two risk bounded trajectories between the given start and goal points as shown in Figure \ref{fig_illus3_1}-a and \ref{fig_illus3_1}-b. We also use RRT-SOS algorithm and time-varying SOS optimization to obtain piece-wise linear trajectories as shown in Figure \ref{fig_illus3_1}-c and \ref{fig_illus3_1}-d, respectively. We note that the standard SOS optimization-based method results in conservative trajectories as shown in Figure \ref{fig_illus3_1}-a and \ref{fig_illus3_1}-b. 

We compare our proposed methods with Monte Carlo-based risk bounded RRT algorithm that uses uncertainty samples and time discretization to look for the risk bounded trajectories. To verify the safety of the edges in the RRT, we use only $10$ uncertainty samples and $20$ uniformly sampled way-points on the edges. Such RRT algorithm is significantly slower and also does not provide any guaranteed risk bounded trajectories.

The run-time for the standard SOS optimization to obtain the piece-wise linear and polynomial trajectories are both roughly $1.5$ seconds. Also, the run-time for the time-varying SOS optimization is roughly $2.2$ seconds (for more information see Section \ref{sec_discuss}). The run-time for the RRT-SOS algorithm to obtain feasible and optimal trajectories are roughly $9.3$ and $92.1$ seconds, respectively. 
The run-time for the Monte Carlo-based risk bounded RRT algorithm to obtain feasible and optimal trajectories are roughly $215.12$ and $3471.4$ seconds, respectively. Also, continuous-time safety verification of each edge in the RRT-SOS algorithm via \eqref{SOS_Cond} takes roughly $0.1$ seconds while the sampling-based safety verification in the Monte Carlo-based risk bounded RRT algorithm takes roughly $3$ seconds.

\begin{figure}[t]
	\centering
	\includegraphics[scale=0.24]{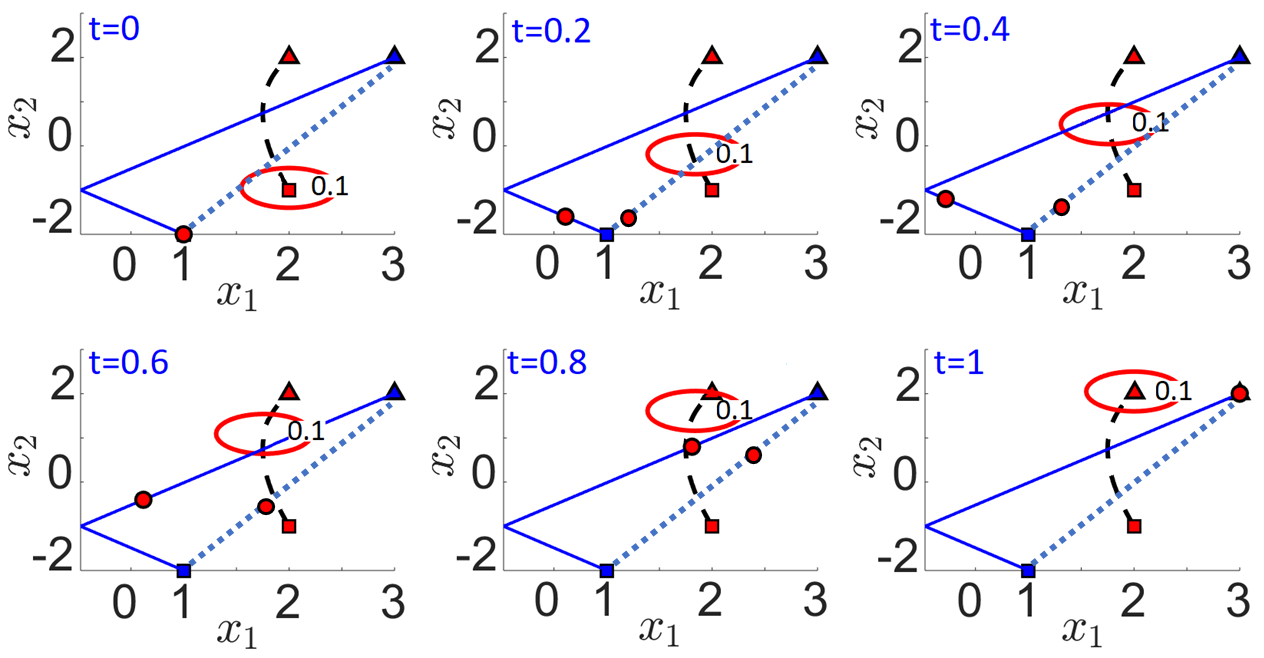}
	\caption{{Illustrative Example 4: Risk bounded trajectories between the start (square) and goal (triangle) points obtained via the time-varying SOS optimization (solid line) and RRT-SOS algorithm (dashed line) in the presence of the probabilistic moving obstacle.}}
	\label{fig_illus4_1}
\end{figure}

\subsubsection{Comparison with Linear Gaussian Method}
We compare the RRT-SOS algorithm with the linear Gaussian method proposed in \cite{blackmore2011chance}. 
In their setting, the system dynamics is discrete-time linear time invariant system with Gaussian noise. 
The obstacles are represented by convex polytopes. 
In \Cref{chap4_comparison_lingaussian_1}, we have compared the risk contour of the linear Gaussian method with our method. 
Since the trajectory planning problem does not consider system dynamics, we modify the linear Gaussian method to exclude the system dynamics. 
The modified linear Gaussian method is to solve the following optimization problem:
\begin{align}
& \underset{\mathbf{x}(t): [t_0, t_f] \rightarrow \mathbb{R}^{n_x}}{\text{minimize}}
 \int_{t_0}^{t_f}  \left\Vert \dot{\mathbf{x}}(t) \right\Vert_2^2 dt \label{opt_lin_gaussian_obj} \\
& \text{subject to} \ 
\ \ \mathbf{x}(t_0)=\mathbf{x}_0, \  \mathbf{x}(t_f)=\mathbf{x}_f \tag{\ref{opt_lin_gaussian_obj}a} \\
&  \mathbf{x}(t) \in \widetilde{\mathcal{C}}^{\Delta}_{r_i}   , \ \forall t\in [t_0,t_f], \  i=1,...,n_{o_s} \tag{\ref{opt_lin_gaussian_obj}b} 
\end{align}
Since for each obstacle a risk contour is represented as a union of half planes
\begin{align}
    \widetilde{\mathcal{C}}_r^{\Delta} = \bigcup_{i=1}^\ell \{\tbx \in \mathcal{X}: \mathbf{a}_i^\top \tbx \leq b_i \}, \tag{\ref{obstacle_lin_gaussian}}
\end{align}
if we look for piecewise linear trajectories \eqref{linear}, then the optimization problem \eqref{opt_lin_gaussian_obj} becomes a disjunctive programming problem, which can also be formulated as a mixed integer programming problem.

\begin{figure}
    \centering
    \includegraphics[scale=0.23]{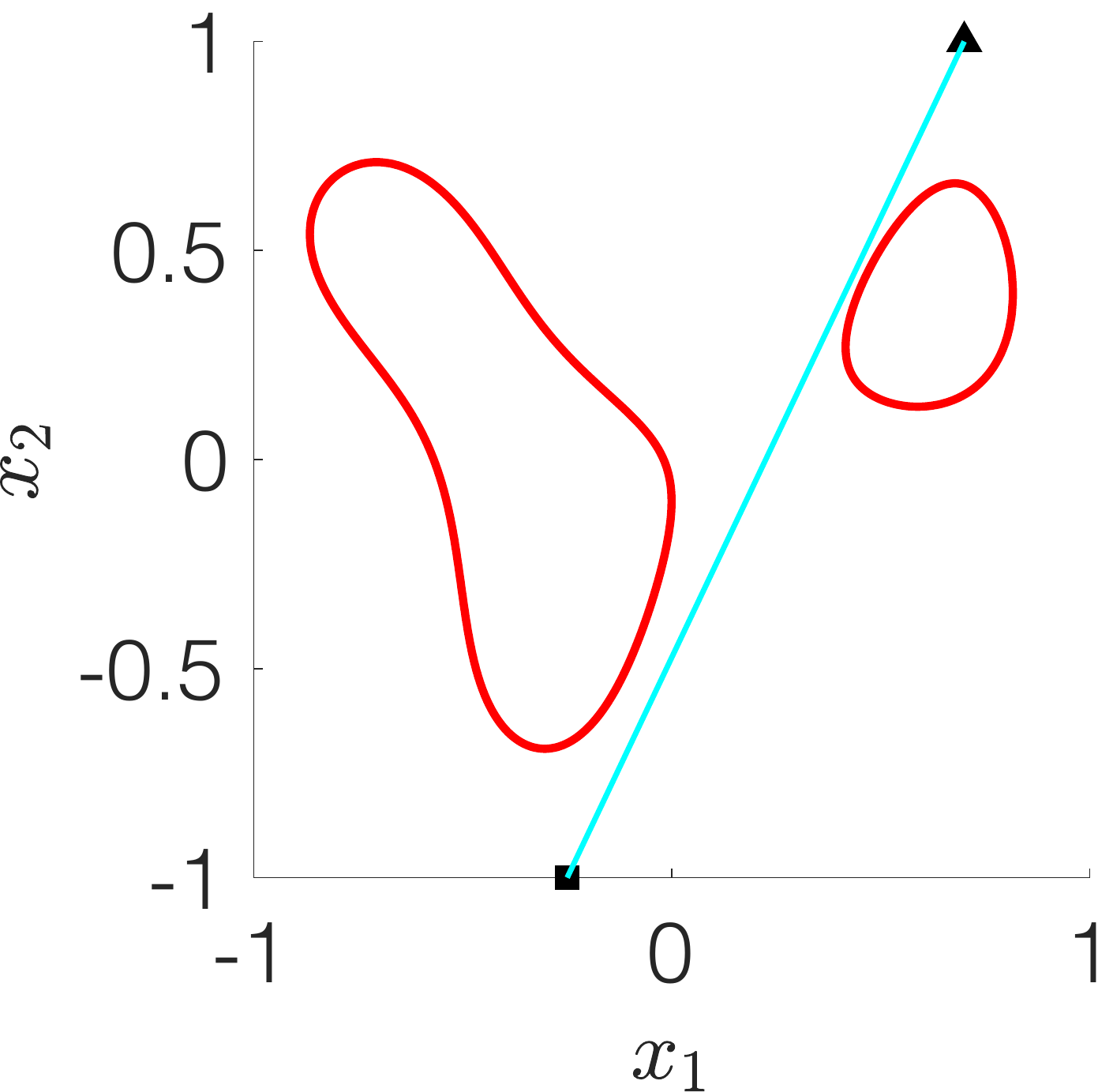}
    \includegraphics[scale=0.23]{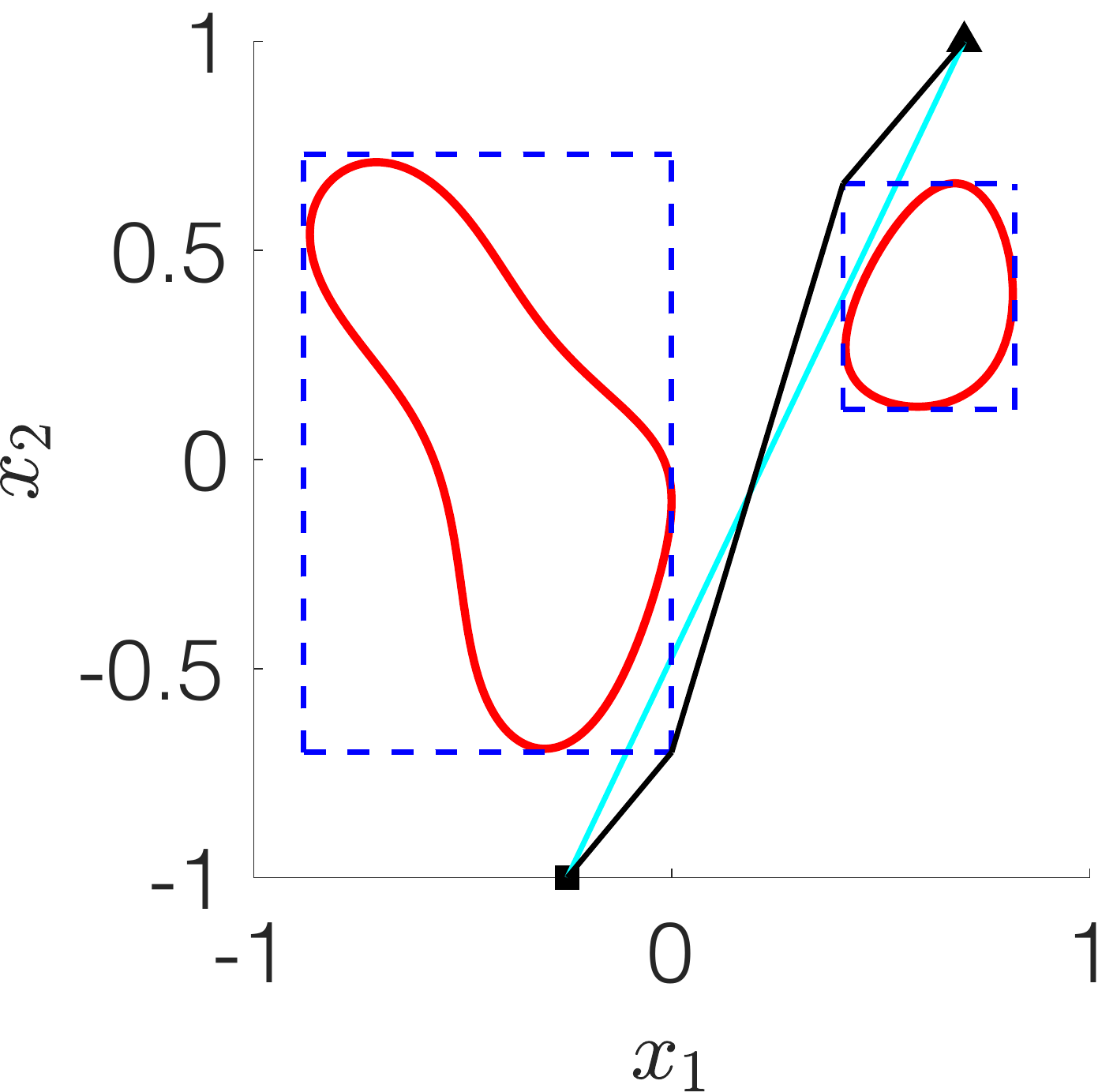}
    \caption{Comparison between RRT-SOS and the modified linear Gaussian method. The square dot represents the initial state and the triangle dot represents the goal state. RRT-SOS is using the risk contour whose boundary is represented by red curves. The cyan line between the initial and the goal states is the path planned by RRT-SOS. In contrast, the blue dotted rectangles are the boundary of the risk contour used by the linear Gaussian method, and the black line segments is the path planned by the linear Gaussian method.}
    \label{fig:chp5_cmp_lin_gaussian}
\end{figure}
Consider the example in \Cref{sec:exp_a1}, the 2D uncertain obstacle represented by a 5th order polynomial. 
We want to plan a trajectory from the initial state $(-0.25,-1)$ to the goal state $(0.7,1)$ with bounded risk of 0.1. 
Using the risk contour obtained from our analytical method, the RRT-SOS algorithm finds the risk bounded path as a straight line between the initial state and the goal state, plotted as a cyan line in \Cref{fig:chp5_cmp_lin_gaussian}. 
The linear Gaussian method uses a more conservative obstacle representation and finds the piecewise linear path consisting of 3 line segments, plotted as black line segments in \Cref{fig:chp5_cmp_lin_gaussian}. 

The advantages of our method include: i) Our method utilizes a less conservative risk contour. ii) Our method is more efficient. Our method is solving convex programming programs in building the RRT tree, while the linear Gaussian method is solving a disjunctive programming problem and needs to manually try out certain parameters such as the number of waypoints in the trajectory. 
When the number of obstacles is large, the disjunctive programming problem becomes hard to solve, while our method checks one obstacle at a time using a convex program and hence the complexity of the convex program is not affected by the number of obstacles. iii) Our method provides safety guarantees for the entire continuous-time trajectory, while the linear Gaussian method does not.

\subsection{Planning in Dynamic Uncertain Environments}\label{section:planning_dyn}
In this section, we are concerned with continuous-time risk bounded trajectory planning in the presence of dynamic uncertain obstacles of the form \eqref{obs_d}. More precisely, we aim at solving 
the optimization problem in \eqref{opt_obj} in the presence of dynamic uncertain obstacles $\mathcal{X}_{obs_i}(\omega_i,t), i=1,...,n_{o_d}$, i.e.,

\begin{align} 
& \underset{\mathbf{x}(t): [t_0, t_f] \rightarrow \mathbb{R}^{n_x}}{\text{minimize}}
 \int_{t_0}^{t_f}  \left\Vert \dot{\mathbf{x}}(t) \right\Vert_2^2 dt \label{opt_d_obj} \\
& \text{subject to} \ 
\ \ \mathbf{x}(t_0)=\mathbf{x}_0, \  \mathbf{x}(t_f)=\mathbf{x}_f \tag{\ref{opt_d_obj}a} \\
&  \ \hbox{Prob}\left( \mathbf{x}(t) \in \mathcal{X}_{obs_i}(\omega_i,t)    \right) \leq \Delta, \ \forall t\in [t_0,t_f] \ |_{i=n_{o_s}+1}^{n_{o_d}} \tag{\ref{opt_d_obj}b} \label{opt_d_prob1} 
\end{align}
To obtain the deterministic optimization, we replace probabilistic constraints \eqref{opt_d_prob1} with deterministic constraints in terms of the dynamic $\Delta$-risk contours as follows:

\begin{align} 
& \underset{\mathbf{x}(t): [t_0, t_f] \rightarrow \mathbb{R}^{n_x}}{\text{minimize}}
 \int_{t_0}^{t_f}  \left\Vert \dot{\mathbf{x}}(t) \right\Vert_2^2 dt \label{opt_d2_obj} \\
& \text{subject to} \ 
\ \ \mathbf{x}(t_0)=\mathbf{x}_0, \  \mathbf{x}(t_f)=\mathbf{x}_f \tag{\ref{opt_d2_obj}a} \\
&  \mathbf{x}(t) \in \hat{\mathcal{C}}^{\Delta}_{r_i}(t)   , \ \forall t\in [t_0,t_f], \ i=n_{o_s}+1,...,n_{o_d} \tag{\ref{opt_d2_obj}b} \label{opt_d2_prob1} 
\end{align}
where $\hat{\mathcal{C}}^{\Delta}_{r_i}(t)$ is the dynamic $\Delta$-risk contour of the dynamic uncertain obstacle $\mathcal{X}_{obs_i}(\omega_i,t)$. The set of $\hat{\mathcal{C}}^{\Delta}_{r_i}(t), i=o_{n_s}+1,...,o_{n_d}$ represents the inner approximation of the feasible set of the probabilistic optimization in \eqref{opt_d_obj}. In the deterministic polynomial optimization of \eqref{opt_d2_obj}, we need to make sure that the constraints are satisfied over the entire planning time horizon $[t_0,t_f]$. To solve the time-varying deterministic polynomial optimization in \eqref{opt_d2_obj}, we will use i) time-varying SOS optimization described in Section \ref{sec_s_plan} and also ii) sampling-based motion planning algorithm similar to the one in Section \ref{sec_RRT} that uses the SOS-based continuous-time safety verification. 
In the sampling-based algorithm, given the dynamic nature of the environment, one needs to verify the safety of the trajectory between the given two samples only for the time interval that is needed to traverse between the points. In this paper, we will use the following naive RRT-SOS algorithm:
RRT algorithm looks for piece-wise linear trajectories with $s$ number of linear pieces defined over the time intervals $\Delta t_i = [t_{i-1},t_i), i=1,...,s$ as in \eqref{linear}. To construct the tree, we first fix the number of linear pieces $s$ and build the tree for each time interval incrementally. To expand the tree for the time interval $\Delta t_i$, we connect the given sample point to a vertex in the tree, built for the time interval $\Delta t_{i-1}$, if the linear trajectory between the two points satisfies SOS condition \eqref{SOS_Cond} for the time interval $\Delta t_i$.
Moreover, to construct the tree for the time interval $\Delta t_{s-1}$, we verify the safety of the linear trajectory between the given sample and the goal point for the time interval $\Delta t_s$ as well.

\textbf{Illustrative Example 4:} Consider the uncertain moving obstacle in illustrative example 2. We want to find a risk bounded trajectory between the points $\mathbf{x}(0)=[1,-2]$ and $\mathbf{x}(1)=[3,2]$ by solving the probabilistic optimization in \eqref{opt_d_obj} with $\Delta=0.1$. For this purpose, we solve the deterministic optimization problem in \eqref{opt_d2_obj} with respect to the dynamic $0.1$-risk contour of the uncertain obstacle via the time-varying SOS optimization and RRT-SOS algorithm as shown in Figure \ref{fig_illus4_1}. Note that although the obtained RRT-SOS trajectory looks like a straight line, it is a piece-wise linear trajectory consisting of 2 pieces with different velocities. The run-time for the time-varying SOS optimization and RRT-SOS algorithm are roughly $6.9$ and $0.5$ seconds, respectively.

\section{Continuous-time risk bounded tube design using risk contours} \label{sec_tube}
In this section, we extend planning continuous-time risk bounded trajectories to planning continuous-time risk bounded tubes in the presence of both static and dynamic uncertain obstacles. 
As in trajectory planning, we convert the probabilistic constraints into deterministic constraints as follows:
\begin{align} 
& \underset{\substack{\mathbf{x}(t): [t_0, t_f] \rightarrow \mathbb{R}^{n_x}\\ Q(t) \succ 0, t\in [t_0, t_f]}}{\text{minimize}}
 \int_{t_0}^{t_f}  \left\Vert \dot{\mathbf{x}}(t) \right\Vert_2^2 dt + w\max_{t\in[t_0,t_f]} \lambda_{max}(Q(t))\label{opt2_det_obj} \\
& \text{subject to}  \ 
\ \ \mathbf{x}(t_0)=\mathbf{x}_0, \  \mathbf{x}(t_f)=\mathbf{x}_f \tag{\ref{opt2_det_obj}a} \\
&  \ \mathcal{Q}({\tbx}(t),t) \subseteq \hat{\mathcal{C}}^{\Delta}_{r_i}, \ \forall t\in [t_0,t_f] \ |_{i=1}^{n_{o_s}} \tag{\ref{opt2_det_obj}b} \label{opt2_det_prob1} \\
& \  \mathcal{Q}({\tbx}(t),t) \subseteq \hat{\mathcal{C}}^{\Delta}_{r_i}(t),\  \forall t\in [t_0,t_f] \ |_{i=n_{o_s}+1}^{n_{o_d}} \tag{\ref{opt2_det_obj}c} \label{opt2_det_prob2}\\
& \ \mathcal{Q}({\tbx}(t),t) \subseteq M(t), \forall t\in [t_0,t_f] \tag{\ref{opt2_det_obj}d}
\end{align}
where $\hat{\mathcal{C}}^{\Delta}_{r_i}$ and $\hat{\mathcal{C}}^{\Delta}_{r_i}(t)$ are $\Delta$-risk contours of static and dynamic obstacles, representing inner approximations of the feasible set of the original probabilistic optimization in \eqref{opt2_obj}.

There are two tube design choices. \textbf{Method (1)}: The first choice is that we use the methods in the previous section to plan a polynomial trajectory or a piecewise linear trajectory, $\bar{\tbx}(t)$. Then we build a max-sized tube $\mathcal{Q}(\bar{\tbx}(t),t)$ along this trajectory.

\begin{algorithm}[t]
\caption{Max-sized tube algorithm}
\begin{algorithmic}[1]
\Procedure{BuildTube}{$\tbx(t),I_t,c_{max},\varepsilon$}       
    \State $l \leftarrow 0$
    \State $r \leftarrow c_{max}$
    \If{Verify($\tbx(t),I_t,l$)}
        \While{$r-l>\varepsilon$}
            \State $m \leftarrow (l+r)/2$
            \If{Verify($\tbx(t),I_t,m$)}
                \State $l \leftarrow m$
            \Else
                \State $r \leftarrow m$
            \EndIf 
        \EndWhile
        \State Return Size = $l$
    \Else
        \State Return FAIL
    \EndIf
\EndProcedure

\end{algorithmic}\label{alg:buildtube}
\end{algorithm}

To search for the max-sized tube with respect to parameterization between two points, we use Algorithm \ref{alg:buildtube}, a binary search algorithm that uses verification of tube safety as a sub-procedure. The algorithm searches in a one-dimensional parameter space that controls the size of the tube. If the size of the tube is actually controlled by multiple parameters, the other parameters can be temporarily fixed, while one parameter is being searched. 
The input to the algorithm is the trajectory $\tbx(t)$, the interval $I_t$ for the parameter $t$, the upperbound $c_{max}$ on the parameter that controls the tube size, and the tolerance $\varepsilon$ within which the difference between the output of the algorithm and the true max tube size should lie.
The parameter $c$ to be searched for has a lowerbound $l=0$ and an upperbound $r=c_{max}$. 
The Verify$(\tbx(t), I_t, c)$ function takes in the trajectory $\tbx(t)$, the interval $I_t$, and the current tube parameter $c$.
It returns True if the tube with parameter $c$ is verified safe, and False if the tube is not verified to be safe.
Every iteration, we consider the middle point $m$ of the interval $[l,r]$. 
If the tube with parameter $m$ is verified to be safe, then the optimal $c$ is in the interval $[m,r]$, otherwise it is in the interval $[l,m]$. 
Once the length of the search interval $[l,r]$ is within $\varepsilon$, the algorithm output the obtained parameter $c_{out} = l$. Since the optimal parameter $c_{opt}$ is in $[l,r]$, the difference between $c_{out}$ and $c_{opt}$ is $\leq \varepsilon$.

The most important sub-procedure for the algorithm is to verify if the tube with parameter $c$ is safe. As in \eqref{SOS_Cond}, we can verify the safety of the tube using SOS programming \cite{jasour2021real}. 
Let $\mathcal{S}(t)=\{\mathbf{x}: g_i(\tbx,t)\geq 0, i=1,...,\ell \}$ be the feasible set of optimization \eqref{opt2_det_obj} at time $t$, $t\in [t_0,t_f]$, i.e., the set constructed by the polynomial constraints of all the risk contours $\hat{\mathcal{C}}^{\Delta}_{r_i}, i=1,...,n_{o_s}$, $\hat{\mathcal{C}}^{\Delta}_{r_j}(t), j=n_{o_s}+1,...,n_{o_d}$, as well as the upperbound constraint $M(t)$. 
Let $\mathbf{x}(t)=\sum_{\alpha=0}^d \mathbf{c}_{\alpha}t^{\alpha}, t\in[t_1,t_2]$, where $t_0 \leq t_1\leq t_2 \leq t_f$, be the given trajectory between the two samples $\mathbf{x}_1 \in \mathcal{X}$ and $\mathbf{x}_2 \in \mathcal{X}$ in the uncertain environment.
Let $H(t) = \{\tbx: h(\tbx,t) \geq 0\}, t\in [t_1,t_2]$, where $h$ is a polynomial and for any $t\in [t_1,t_2]$ the set $H(t)$ is a compact set. 
Note that $\mathcal{Q}(\tbx(t),t)$ in the form of \eqref{eq:tube_form} is a special case of $H(t)$. Then the following result holds true:

$H(t)$ satisfies the safety constraints of the deterministic optimization in \eqref{opt2_det_obj} over the time interval $[t_1,t_2]$, i.e., $H(t) \subseteq \mathcal{S}(t)$ for all $t\in[t_1,t_2]$, if polynomials $g_i(\mathbf{x},t), i=1,...,\ell,$ take the following SOS representation:
\begin{align}\label{SOS_Cond_tube}
    g_i(\tbx, t) &={\sigma_0}_i(\tbx,t)+ {\sigma_1}_i(\tbx,t)(t-t_1) \nonumber \\
    & \quad  + {\sigma_2}_i(\tbx,t)(t_2-t) +{\sigma_3}_i(\tbx,t)h(\tbx,t)
\end{align}
where ${\sigma_0}_i(\tbx,t), {\sigma_1}_i(\tbx,t), {\sigma_2}_i(\tbx,t),{\sigma_3}_i(\tbx,t), i=1,...,\ell,$ are SOS polynomials with appropriate degrees.

In Algorithm \ref{alg:buildtube}, given a fixed tube parameter $c$, the tube is completed determined, and the corresponding polynomial $h(\mathbf{x},t)$ is formed. 
The Verify function verifies the safety of the tube with parameter $c$ via \Cref{SOS_Cond_tube}. 
Note that SOS programming for verifying the safety of the tube is a sufficient condition but not a necessary condition. 
It is possible that a tube is safe but is not verified to be safe via SOS programming. 
If the tube is verified to be safe, then it must be safe.

\textbf{Method (2)}: The second design choice is that we build a trajectory and a tube around the trajectory simultaneously. We use sampling-based motion planning algorithms similar to RRT-SOS in Section \ref{sec_RRT} and the one in Section \ref{section:planning_dyn}. 
When building the RRT tree, any two samples are first connected by a trajectory, and then instead of verifying the safety of the trajectory between two samples, we verify the safety of tubes containing the trajectory between two samples. However, it can be quite expensive to build max-sized tubes between any two samples for the entire RRT tree. So a reasonable design choice is to set a threshold on the minimum required tube size, e.g., the minimum radius $r_{min}$ for balls. Samples that do not admit the minimum tube are rejected. 
After finding a path from the start point to the goal point, we look for max-sized tube along this path. 
In summary, Method (2) works as follows: When building the RRT tree, a tube of size $r_{min}$ along the edge between the new node and its nearest tree node is checked. 
If the tube is verified safe, then the new node with the edge is added to the RRT tree. 
Otherwise, the new node is discarded. 
Once we have found a path in the RRT tree from the start point to the goal point, we use Algorithm \ref{alg:buildtube} to search for the max-sized tube along the path. 
Therefore, the final tube obtained by Method (2) has radius at least as large as $r_{min}$ on every point along the path from the start point to the goal point. 
This is particularly useful if we do planning directly in the workspace and the robot can be outer approximated by the ball of radius $r_{min}$. 

In Example \ref{sec:exp_c}, we use Method (1) to directly build the tube on top of a pre-planned trajectory. In Example \ref{sec:exp_d}, we use Method (2) to build the underlying trajectory and the tube together, satisfying the minimum tube size requirement. 

\begin{figure}[t]
    \centering
    \includegraphics[width=0.15\textwidth]{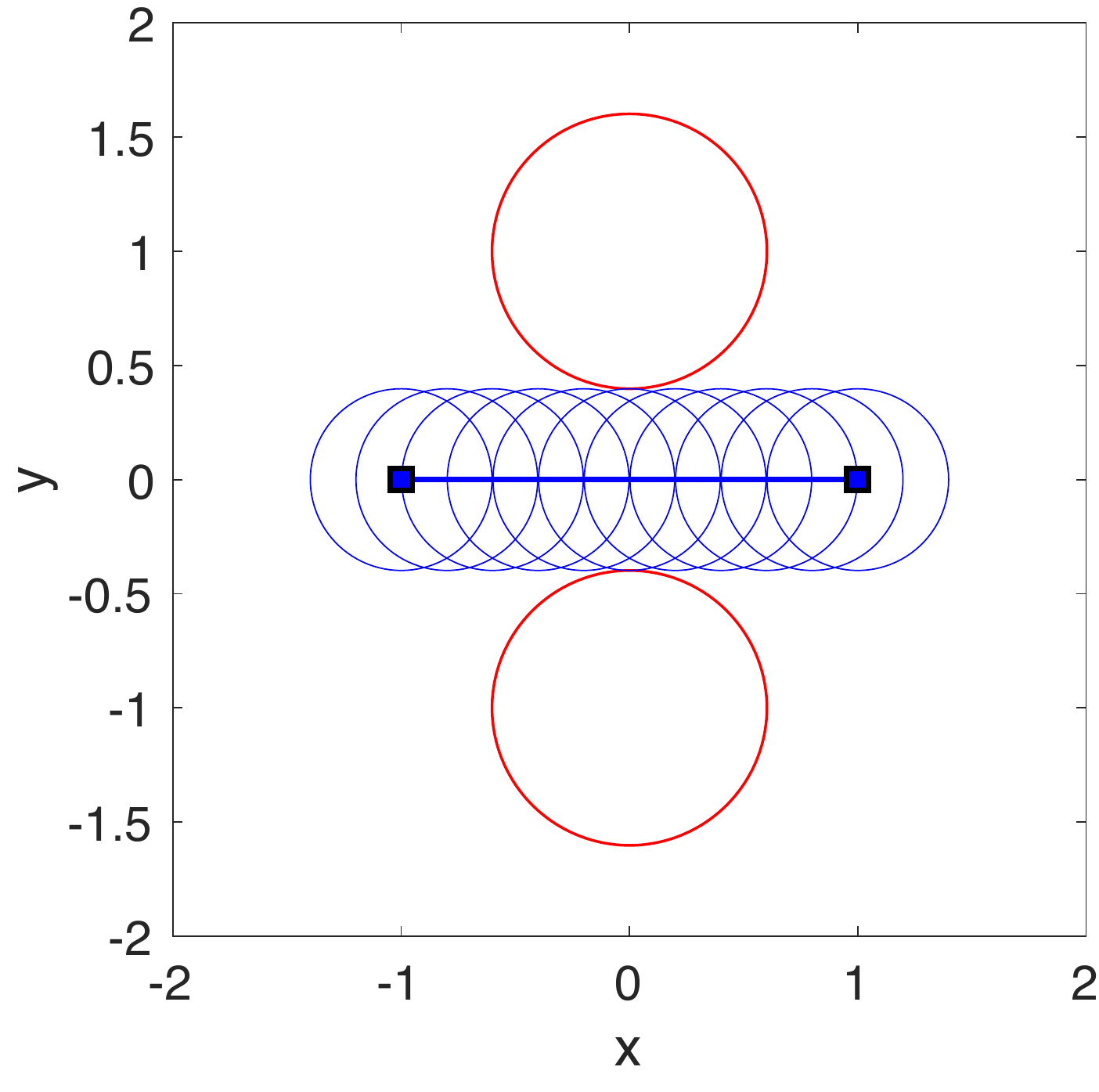}
    \includegraphics[width=0.15\textwidth]{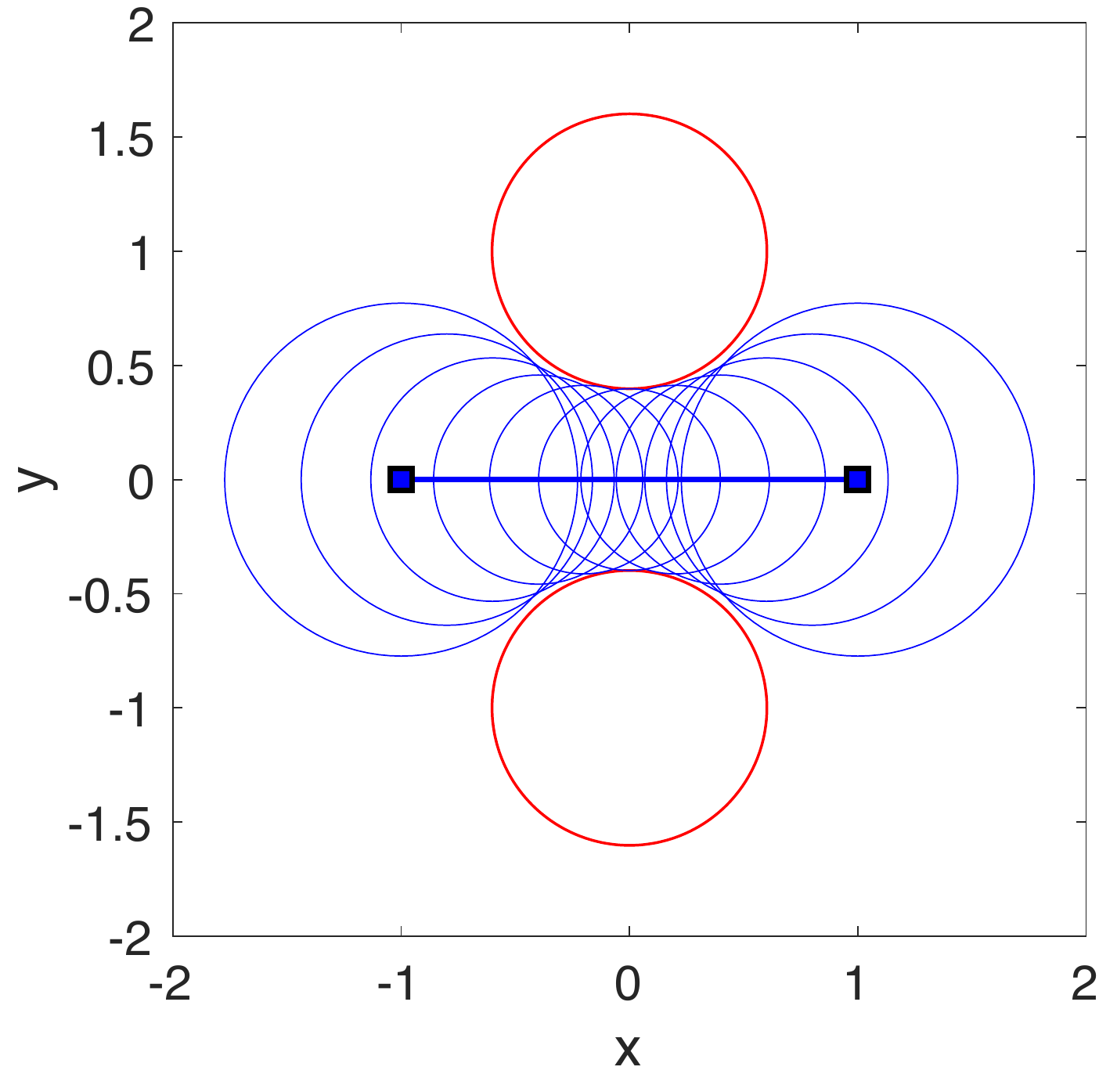}
    \includegraphics[width=0.15\textwidth]{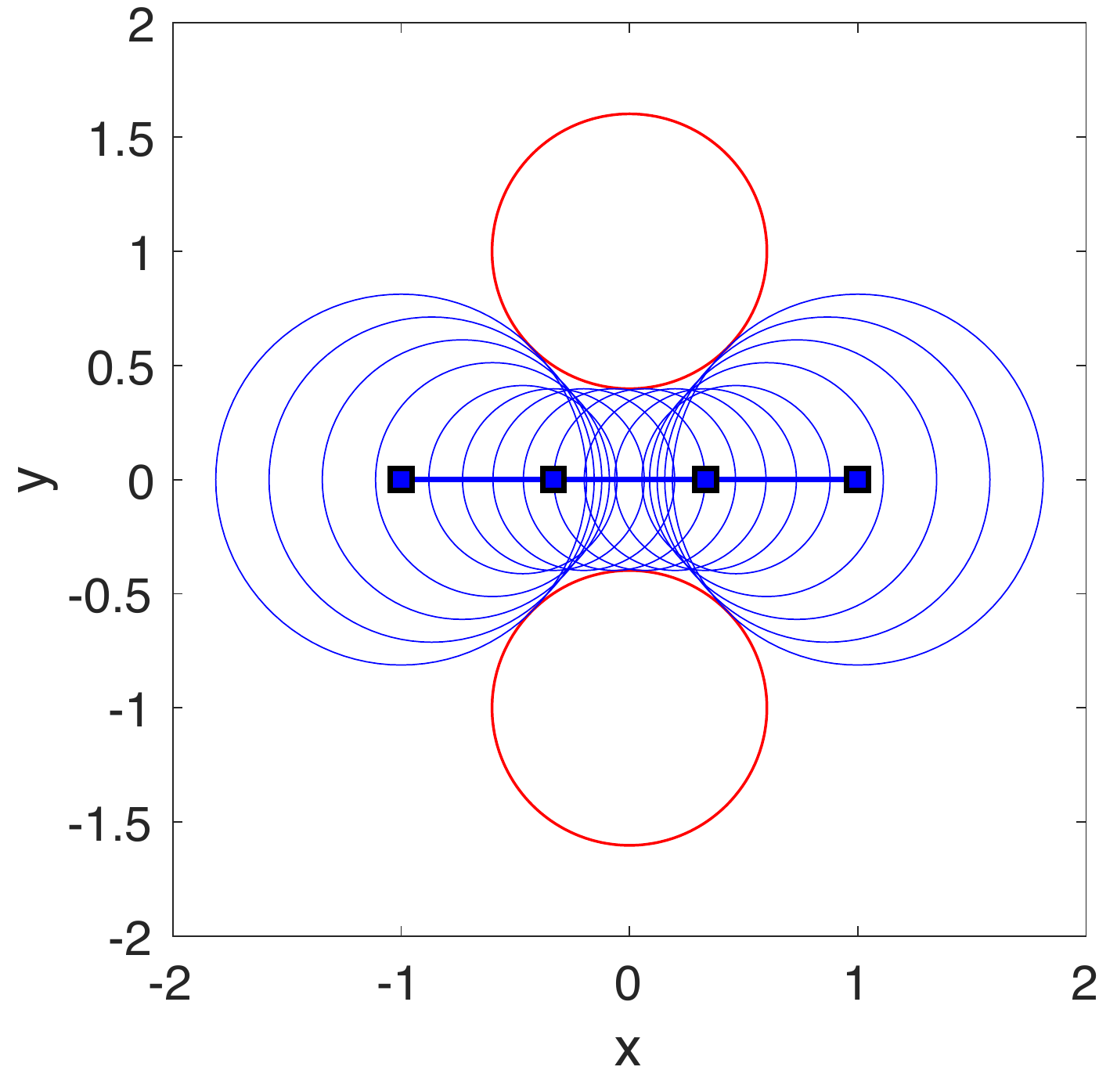}
    \caption{Illustrative Example 5: Left: Constant-sized tube. Middle: Quadratic time-varying-sized tube. Right: Piecewise linear time-varying-sized tube.}
    \label{fig:exp1_tube}
\end{figure}

\textbf{Illustrative Example 5:} Consider two round obstacles with radius 0.5, and their centers at $(0,\pm 1)$ on average with Gaussian noise $N(\mu=0,\sigma^2 = 0.001)$ in both $x$ and $y$ direction. \Cref{fig:exp1_tube} plots the 0.1-risk contours in red. We are going to build a tube between the points $\tbx(0) = (-1,0)$ and $\tbx(1) = (1,0)$, and the underlying trajectory $\bar{\tbx}(t) = \tbx(0) + (\tbx(1) - \tbx(0))t = (-1 + 2t,0)$ for $t\in [0,1]$. We can consider (1) constant-sized tube, (2) quadratic time-varying-sized tube, and (3) piecewise linaer time-varying-sized tube. 

The constant-sized tube is parameterized as $\mathcal{Q}(\bar{\tbx}(t), t) = \{\tbx: c^2 - ||\tbx-\bar{\tbx}(t)||_2^2 \geq 0 \}, t\in [0,1]$. 
We search the parameter $c$ using Algorithm \ref{alg:buildtube}, and the maximum $c = 0.3980$. 

The quadratic time-varying-sized tube is parameterized as $\mathcal{Q}(\bar{\tbx}(t), t) = \{\tbx: (a(t-b)^2+c)^2 - ||\tbx-\bar{\tbx}(t)||_2^2 \geq 0, a>0,0\leq b\leq 1,c\geq 0 \}$, $t\in [0,1]$, where the radius $r(t) = a(t-b)^2+c$ is a quadratic function in $t$. The parameter $a>0$ indicates the change rate of the radius. The larger $a$ is, the faster the radius changes over $t \in [0,1]$. The parameter $b \in [0,1]$ represents the axis of symmetry of the parabola. $r(t)$ decreases over $t\in [0,b]$ and increases over $t\in [b,1]$. The parameter $c$ is the value of the minimum point on the parabola. In this example, we fix $a = 1.5, b = 0.5$, and use Algorithm \ref{alg:buildtube} to search for the parameter $c$. We get the maximum $c = 0.3980$. The quadratic time-varying-sized tube covers more safe regions than the constant sized tube.

Finally we consider piecewise linear time-varying-sized tube. We break the line segment between $(-1,0)$ and $(1,0)$ into three smaller line segments, by add two waypoints at $(-1/3,0)$ and $(1/3,0)$. On the line segment between $(1/3,0)$ and $(1,0)$, we consider a linear time-varying-sized tube of the form $\mathcal{Q}(\bar{\tbx}(t), t) = \{\tbx: (at+c)^2 - ||\tbx-\bar{\tbx}(t)||_2^2 \geq 0, a>0,c\geq 0 \}$, $t\in [0,1]$. The radius $r(t)=at+c$ is linear in $t$. We fix $a = 0.5$ and use Algorithm \ref{alg:buildtube} to search for the parameter $c$ and get the maximum $c = 0.3122$. Similarly, on the line segment between $(-1,0)$ and $(-1/3,0)$, we consider a linear time-varying-sized tube of the form $\mathcal{Q}(\bar{\tbx}(t), t) = \{\tbx: (a(1-t)+c)^2 - ||\tbx-\bar{\tbx}(t)||_2^2 \geq 0, a>0,c\geq 0 \}$, $t\in [0,1]$. We fix $a = 0.5$ and by symmetry get maximum $c = 0.3122$. On the line segment between $(-1/3,0)$ and $(1/3,0)$, the linear time-varying-sized tube degenerates to constant sized tube and the radius is $c = 0.3980$. Near the endpoints $(-1,0)$ and $(1,0)$ the piecewise linear time-varying-sized tube covers more safety space than the quadratic time-varying-sized tube, while near the waypoints $(-1/3,0)$ and $(1/3,0)$, the quadratic time-varying-sized tube covers more safety space.

This example shows that different parametrizations of the tube lead to different max-sized tubes.
There is a trade-off between the complexity of the tube parametrization and the safe region the tube covers. 
The constant-sized tube has the simplest parametrization and it covers the smallest safe region.
The quadratic time-varying-sized tube and the piecewise linear time-varying-sized tube have slightly more complicated parametrizations and they cover more safe regions. 

\section{Experiments}\label{sec_exp}
\begin{figure}[t]
    \centering
    \includegraphics[scale=0.29]{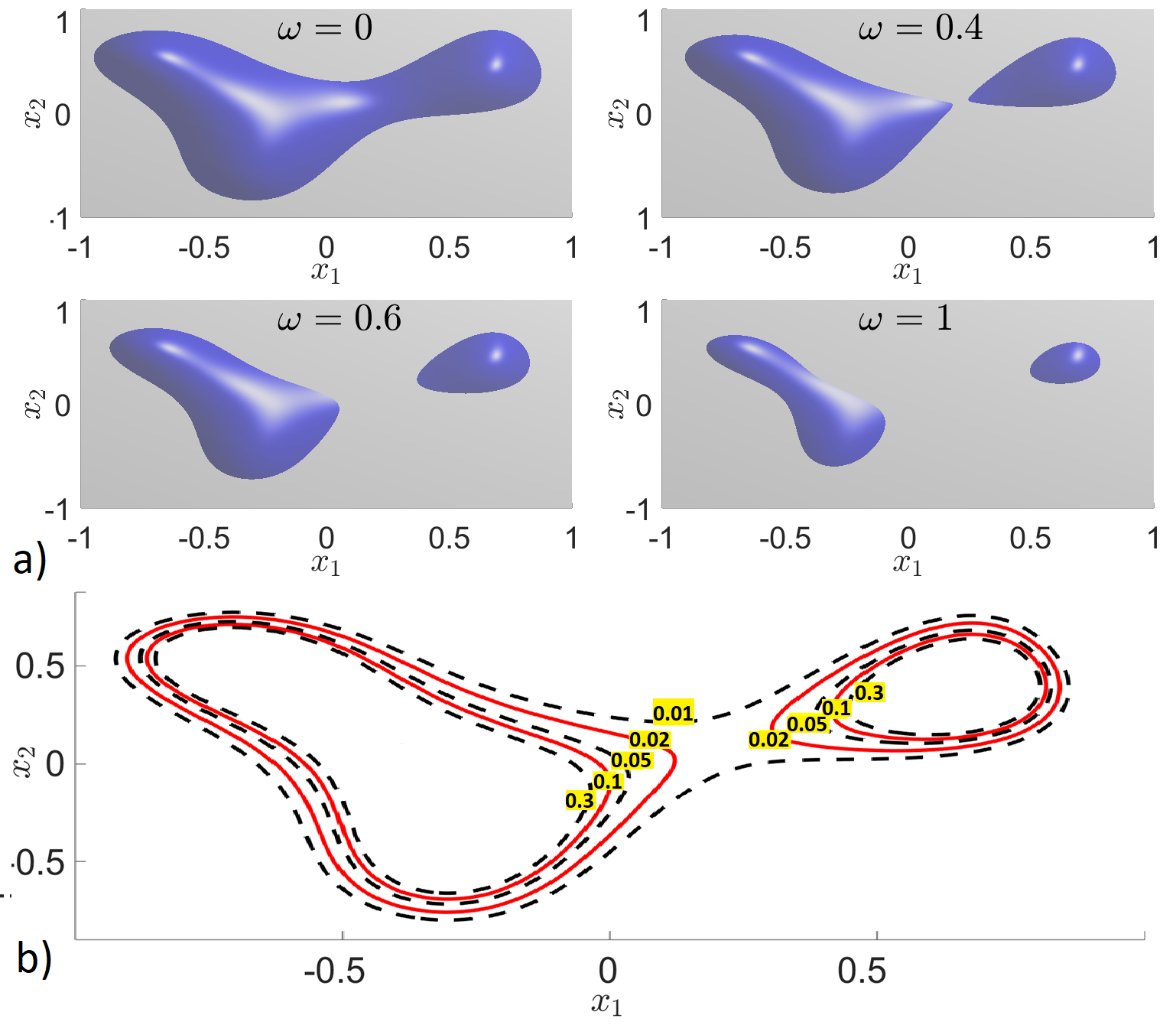}
	\caption{{Example \ref{sec:exp_a1}: a) Nonconvex obstacle with uncertain parameter $\omega \sim Beta(9,0.5)$, b) Obtained static risk contours $\hat{\mathcal{C}}^{\Delta}_r$ defined in \eqref{rc_s_cheby} for different risk levels $\Delta=[0.3,0.1,0.05,0.02,0.01]$.}}
  \label{fig_exa1_1}
\end{figure}

\begin{figure}
	\centering
	\includegraphics[scale=0.29]{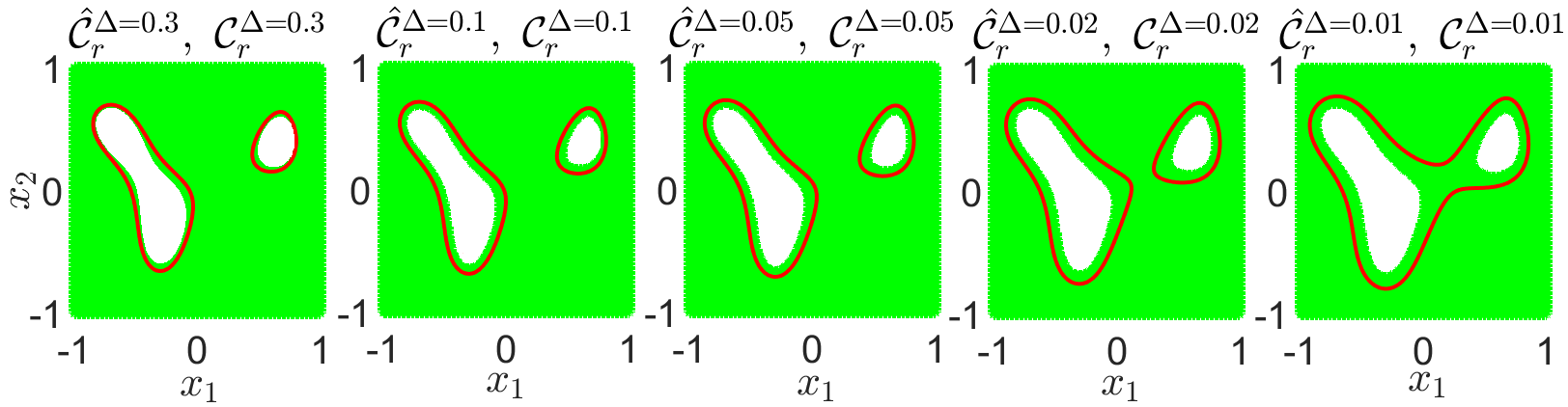}   
	\caption{{Example \ref{sec:exp_a1}: True $\Delta$-risk contour ${\mathcal{C}}^{\Delta}_{r}$ (green) and its inner approximation $\hat{\mathcal{C}}^{\Delta}_{r}$ (outside of the solid-line). Inside the risk contours, probability of collision with the uncertain obstacle is less or equal to the associated risk level $\Delta$.}}
	\label{fig_exa1_2}
\end{figure}

In this section, numerical examples are presented to illustrate the performance of the proposed approaches. To obtain the risk bounded continuous-time trajectories, we use the time-varying SOS optimization and the RRT-SOS algorithm described in Section \ref{sec_plan}. 
Note that the provided RRT-SOS algorithm uses a naive tree search as explained in Section \ref{sec_plan}. 
The main objective of the provided RRT-SOS algorithm is to demonstrate how the risk contours in \eqref{rc_s_cheby} and \eqref{rc_d_cheby} and the SOS-based continuous-time safety constraints in \eqref{SOS_Cond} can be incorporated into sampling-based motion planning algorithms to look for guaranteed risk bounded continuous-time trajectories in stochastic environments. The computations in this section were performed on a computer with Intel i7 2.7 GHz processors and 8 GB RAM. We use the Spotless MATLAB package \cite{Spot_1} to verify the SOS-based continuous-time safety constraints in the RRT-SOS algorithm and the Julia package provided by \cite{SOS_time2} to solve the time-varying SOS optimization.

\subsection{Risk Contours}

The purpose of this example is to demonstrate how the provided approach can be used to compute the risk contours in the presence of highly complex uncertain unsafe regions. 
\subsubsection{2D Uncertain Obstacle}\label{sec:exp_a1}
Static uncertain obstacle of the form \eqref{obs_s} is described by the polynomial $\mathcal{P}(\mathbf{x},\omega)=- 0.42x_1^5 - 1.18x_1^4x_2 - 0.47x_1^4 + 0.3x_1^3x_2^2 - 0.57x_1^3x_2 + 0.6x_1^3 - 0.65x_1^2x_2^3 + 0.17x_1^2x_2^2 + 1.87x_1^2x_2 + 0.06x_1^2 + 0.69x_1x_2^4 - 0.14x_1x_2^3 - 0.85x_1x_2^2 + 0.6x_1x_2 - 0.21x_1 + 0.01x_2^5 - 0.06x_2^4 - 0.07x_2^3 - 0.41x_2^2 - 0.08x_2 +0.07 - 0.1w 
 $ where the uncertain parameter $\omega$ has a Beta distribution with parameters $(9,0.5)$ over $[0,1]$. Figure \ref{fig_exa1_1} shows the uncertain obstacle for different values of the uncertain parameter $\omega$. We obtain the static risk contours defined in \eqref{rc_s_cheby} for different risk levels $\Delta$ as shown in Figure \ref{fig_exa1_1} and \ref{fig_exa1_2}.

\begin{figure}[t]
    \centering
    \includegraphics[scale=0.26]{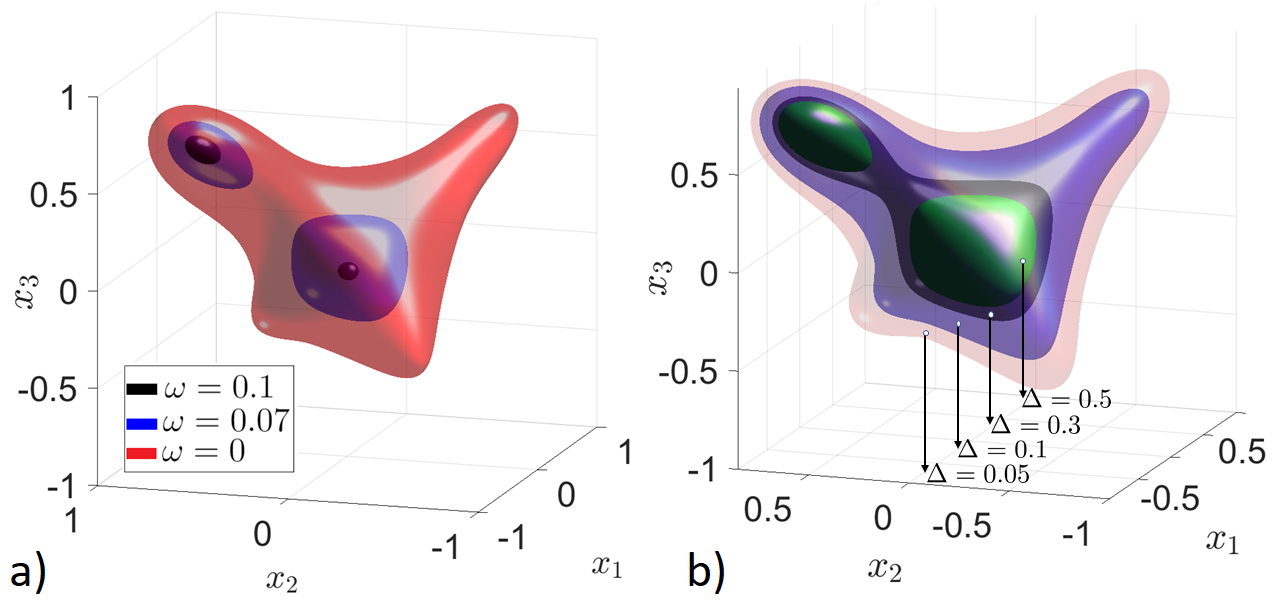}
	\caption{{Example \ref{sec:exp_a2}: a) Nonconvex  obstacle with uncertain parameter $\omega \sim \mathcal{N}(0.1,0.001)$, b) Obtained static risk contours $\hat{\mathcal{C}}^{\Delta}_r$ defined in \eqref{rc_s_cheby} for different risk levels $\Delta=[0.5,0.3,0.1,0.05]$.}}
    \label{fig_exa1_3}
\end{figure}

\subsubsection{3D Uncertain Obstacle}\label{sec:exp_a2}
Static uncertain obstacle of the form \eqref{obs_s} is described by the polynomial $\mathcal{P}(\mathbf{x},\omega)=
0.94-0.002x_1-0.004x_2-0.04x_3-0.38x_1^2+0.04x_1x_2-0.31x_2^2-0.05x_1x_3-0.01x_2x_3-0.4x_3^2-0.1x_1^3-0.02x_1^2x_2+0.09x_1x_2^2-0.05x_2^3+0.14x_1^2x_3-1.83x_1x_2x_3+0.11x_2^2x_3-0.1x_1x_3^2+0.12x_2x_3^2+0.34x_3^3-0.32x_1^4-0.13x_1^3x_2+0.48x_1^2x_2^2+0.11x_1x_2^3-0.34x_2^4+0.03x_1^3x_3+0.01x_1^2x_2x_3-0.005x_1x_2^2x_3-0.05x_2^3x_3+0.54x_1^2x_3^2-0.06x_1x_2x_3^2+0.48x_2^2x_3^2+0.008x_1x_3^3+0.06x_2x_3^3-0.3x_3^4+0.12x_1^5+0.005x_1^4x_2-0.1x_1^3x_2^2+0.007x_1^2x_2^3+0.005x_1x_2^4+0.071x_2^5-0.02x_1^4x_3+0.73x_1^3x_2x_3-0.07x_1^2x_2^2x_3+0.72x_1x_2^3x_3-0.20x_2^4x_3+0.03x_1^3x_3^2-0.01x_1^2x_2x_3^2+0.02x_1x_2^2x_3^2-0.05x_2^3x_3^2-0.07x_1^2x_3^3+0.73x_1x_2x_3^3+0.09x_2^2x_3^3+0.03x_1x_3^4-0.06x_2x_3^4-0.31x_3^5-\omega-0.84
 $ where the uncertain parameter $\omega$ has a normal distribution with mean $0.1$ and variance $0.001$, \cite{Risk_Ind}. Figure \ref{fig_exa1_3} shows the uncertain obstacle for different values of the uncertain parameter $\omega$. We obtain the static risk contours defined in \eqref{rc_s_cheby} for different risk levels $\Delta$ as shown in Figure \ref{fig_exa1_3}.

\begin{figure}[t]
    \centering
    \includegraphics[scale=0.25]{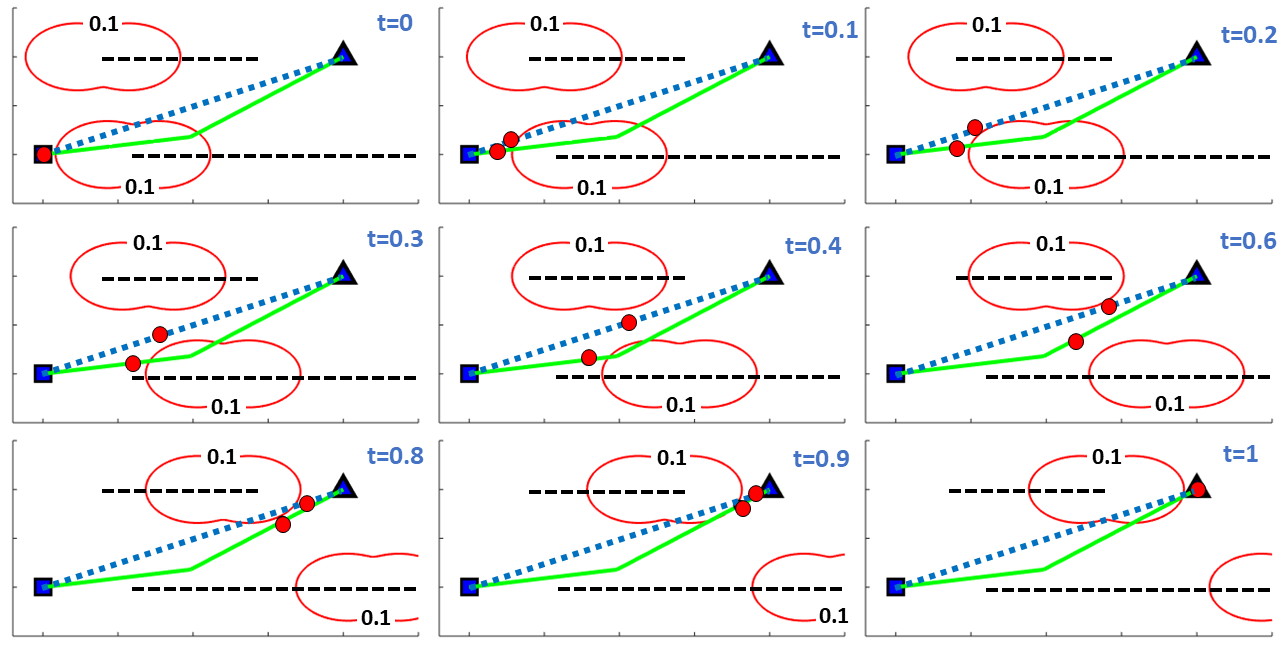}
	\caption{{Example \ref{sec:exp_b}: Risk bounded trajectories for the lane-change maneuver between the start (square) and goal (triangle) points obtained via the time-varying SOS optimization (solid line) and RRT-SOS algorithm (dotted line) in the presence of surrounding vehicles with uncertain trajectories. Dashed lines show the expected values of the uncertain trajectories of the surrounding vehicles. 
	}}
  \label{fig_exa2_2}
\end{figure}


\subsection{Risk Bounded Lane Changing for Autonomous Vehicles} \label{sec:exp_b}
In this example, we generate a risk bounded trajectory for the lane-change maneuver of an autonomous vehicle in the presence of surrounding vehicles. In this scenario, uncertain locations of the surrounding vehicles are modeled as the following sets: 
$\mathcal{X}_{obs_1}(\omega_1,t)=\{(x_1,x_2): 0.3^2 - (x_1-p_{1}(t,\omega_1))^2 - (x_2-1)^2 \geq 0 \}$ and 
$\mathcal{X}_{obs_2}(\omega_2,t)=\{(x_1,x_2): 0.3^2 - (x_1-p_{2}(t,\omega_1))^2 - x_2^2 \geq 0 \}$
where 
$p_{1}(t,\omega_1)=t+0.4+\omega_1$ and $p_{2}(t,\omega_2)=2t+0.6+\omega_2$ are the uncertain trajectories of the surrounding vehicles with uncertain parameters $\omega_i\sim Uniform[-0.1,0.1],i=1,2$. For the lane-change maneuver, we look for a risk bounded trajectory between the points $\mathbf{x}(0)=(0,0)$ and $\mathbf{x}(1)=(2,0)$ over the planning time horizon $t\in[0,1]$. Figure \ref{fig_exa2_2} shows the obtained trajectory using the time-varying SOS optimization and RRT-SOS algorithm considering the dynamic $0.1$-risk contours of the surrounding vehicles. 
The run-time for the time-varying SOS optimization and RRT-SOS are roughly 1.2 and 6.5 seconds, respectively.

\begin{figure}[t]
    \centering
    \includegraphics[scale=0.55]{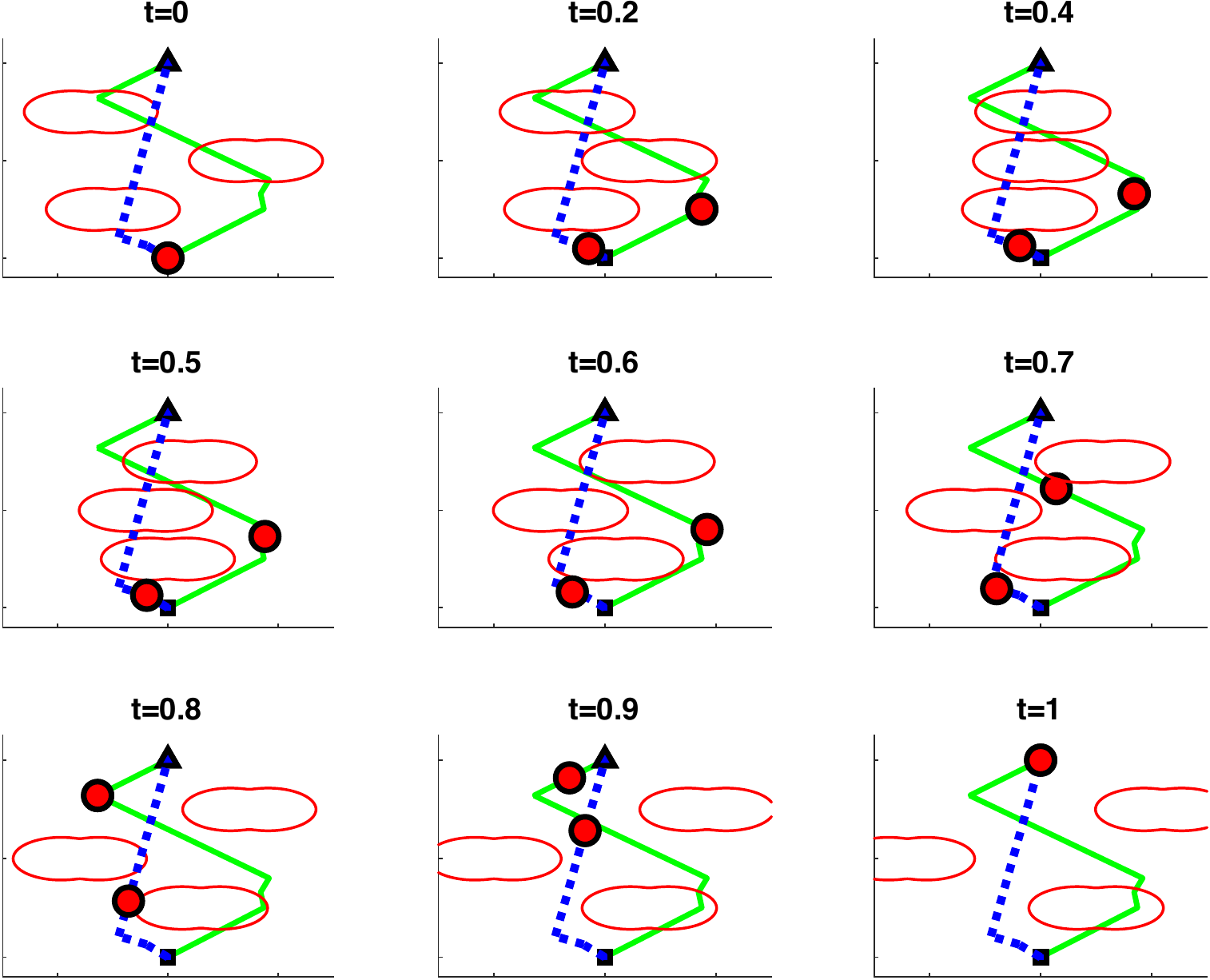}
	\caption{{Example \ref{sec:exp_c1}: Risk bounded trajectories for a delivery robot between the start (square) and goal (triangle) points obtained via the time-varying SOS optimization (solid line) and RRT-SOS algorithm (dotted line) in the presence of moving uncertain obstacles (red curves). 
	}}
  \label{fig_exa3_2}
\end{figure}
\begin{figure}[t]
    \centering
    \includegraphics[scale=0.55]{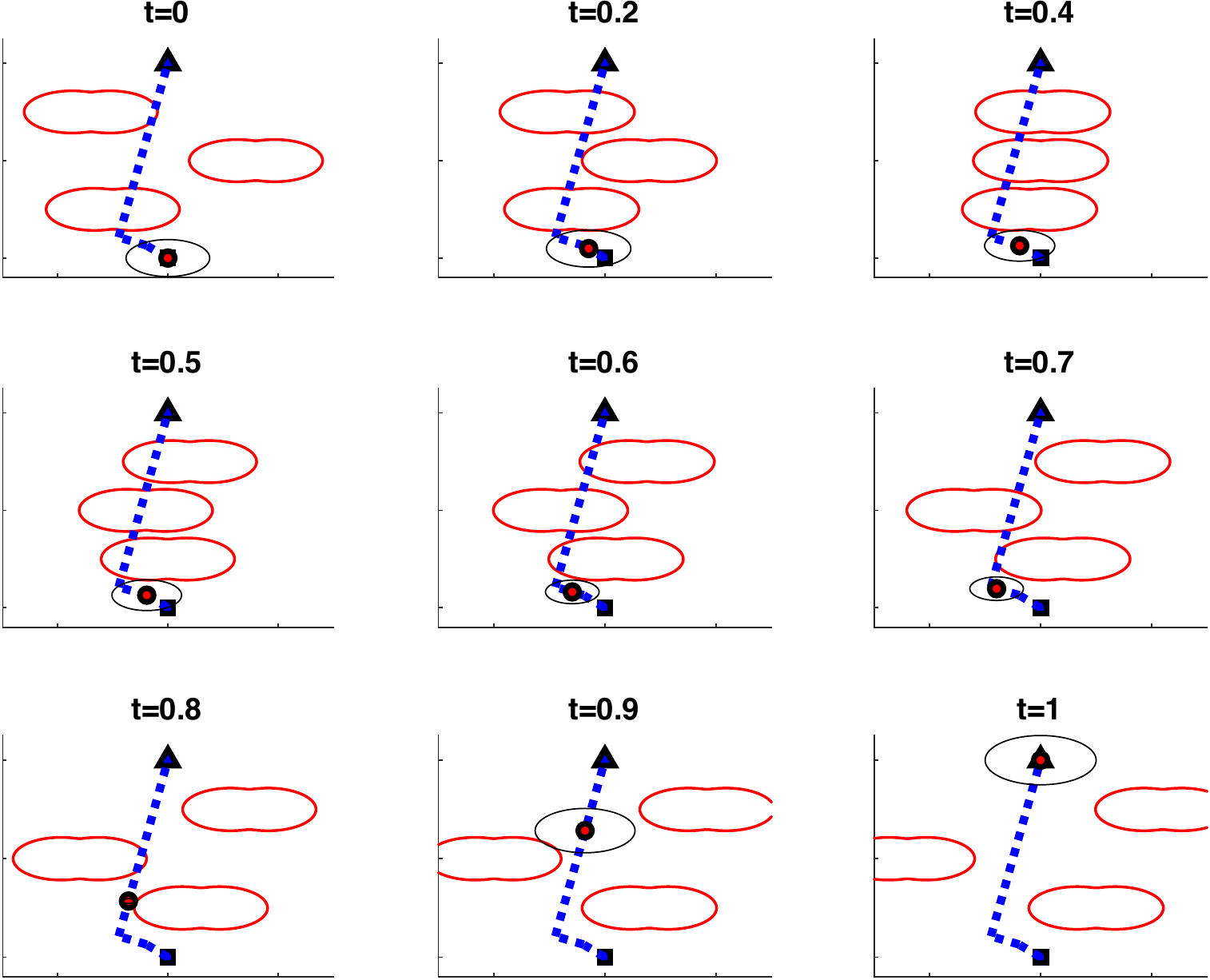}
	\caption{{Example \ref{sec:exp_c2}: The risk bounded tube (black circles) for a delivery robot (red dot) between the start (square) and goal (triangle) points. The tube is directly built on top of the trajectory obtained via RRT-SOS algorithm (dotted line). 
	}}
  \label{fig_exa3_2_2}
\end{figure}

\subsection{Risk Bounded Trajectory Planning and Tube Design for Delivery Robots}\label{sec:exp_c}

\subsubsection{Risk Bounded Trajectory Planning} \label{sec:exp_c1}
In this example, we generate a risk bounded trajectory for a delivery robot in the presence of uncertain moving obstacles. In this scenario, uncertain locations of the moving obstacles are modeled as the following sets: $\mathcal{X}_{obs_1}(\omega_1,t)=\{(x_1,x_2): 0.4^2 - (-x_1+p_{1}(t,\omega_1))^2 - (x_2-1)^2 \geq 0 \}$, $\mathcal{X}_{obs_2}(\omega_2,t)=\{(x_1,x_2): 0.4^2 - (x_1+p_{2}(t,\omega_1))^2 - (x_2-2)^2 \geq 0 \}$, and $\mathcal{X}_{obs_3}(\omega_3,t)=\{(x_1,x_2): 0.4^2 - (-x_1+p_{3}(t,\omega_3))^2 - (x_2-3)^2 \geq 0 \}$ where $p_{1}(t,\omega_1)=t - 0.5+\omega_1$, $p_{2}(t,\omega_1)= 2t-\omega_2-0.8$, and $p_{3}(t,\omega_3)=1.8t - 0.7+\omega_3$ are the uncertain trajectories of the moving obstacles with uncertain parameters $\omega_i\sim Uniform[-0.1,0.1],i=1,2,3$. We look for a risk bounded trajectory between the start and destination points, $\mathbf{x}(0)=(0,0)$ and $\mathbf{x}(1)=(0,4)$, over the planning time horizon $t\in[0,1]$. Figure \ref{fig_exa3_2} shows the obtained trajectories using the time-varying SOS optimization and RRT-SOS algorithm considering the dynamic $0.1$-risk contours of the moving obstacles. 
The run-time for the time-varying SOS optimization and RRT-SOS are roughly 7 and 197 seconds, respectively.

\begin{figure}[t]
    \centering
    \includegraphics[scale=0.45]{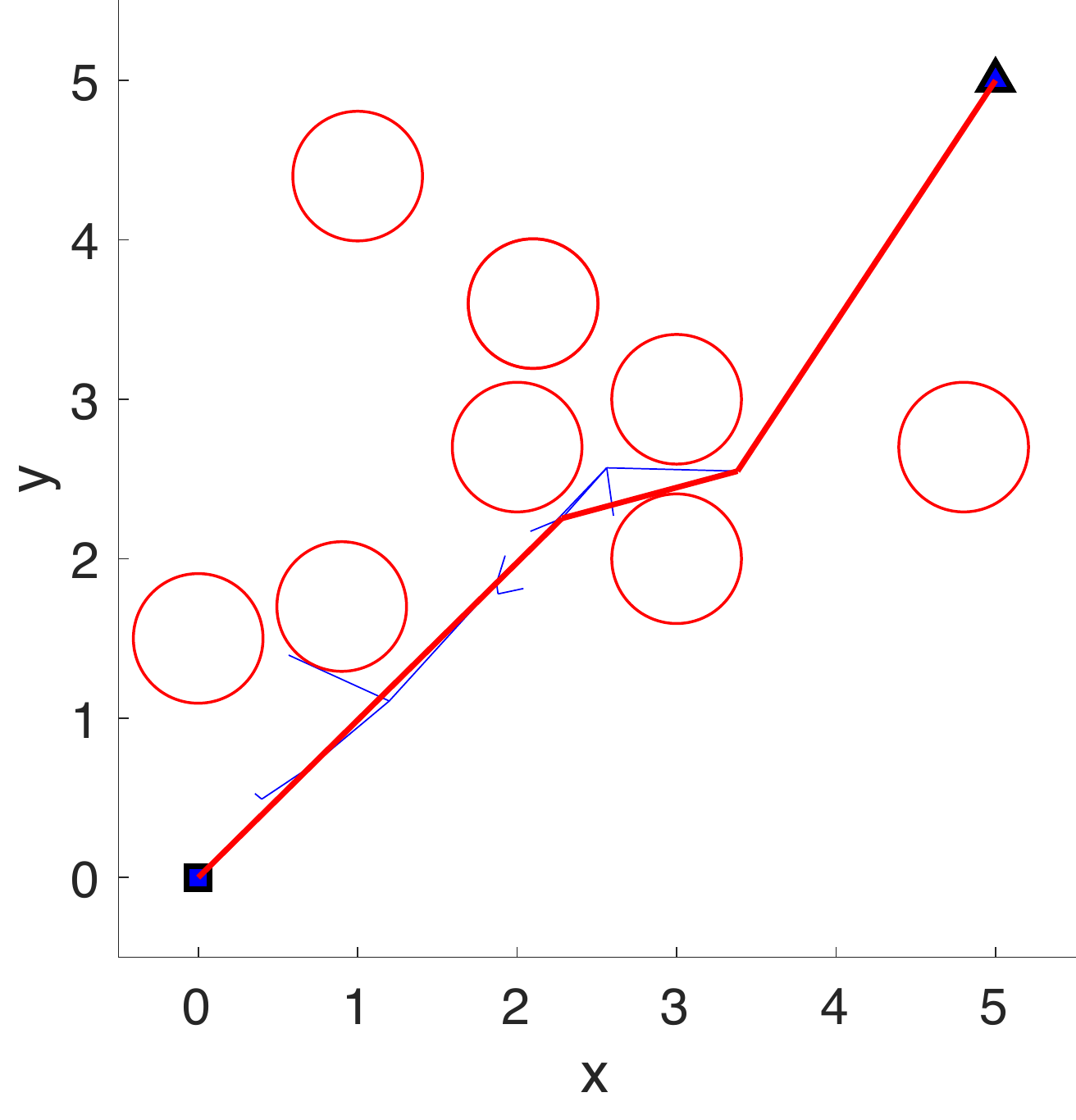}
    \includegraphics[scale=0.45]{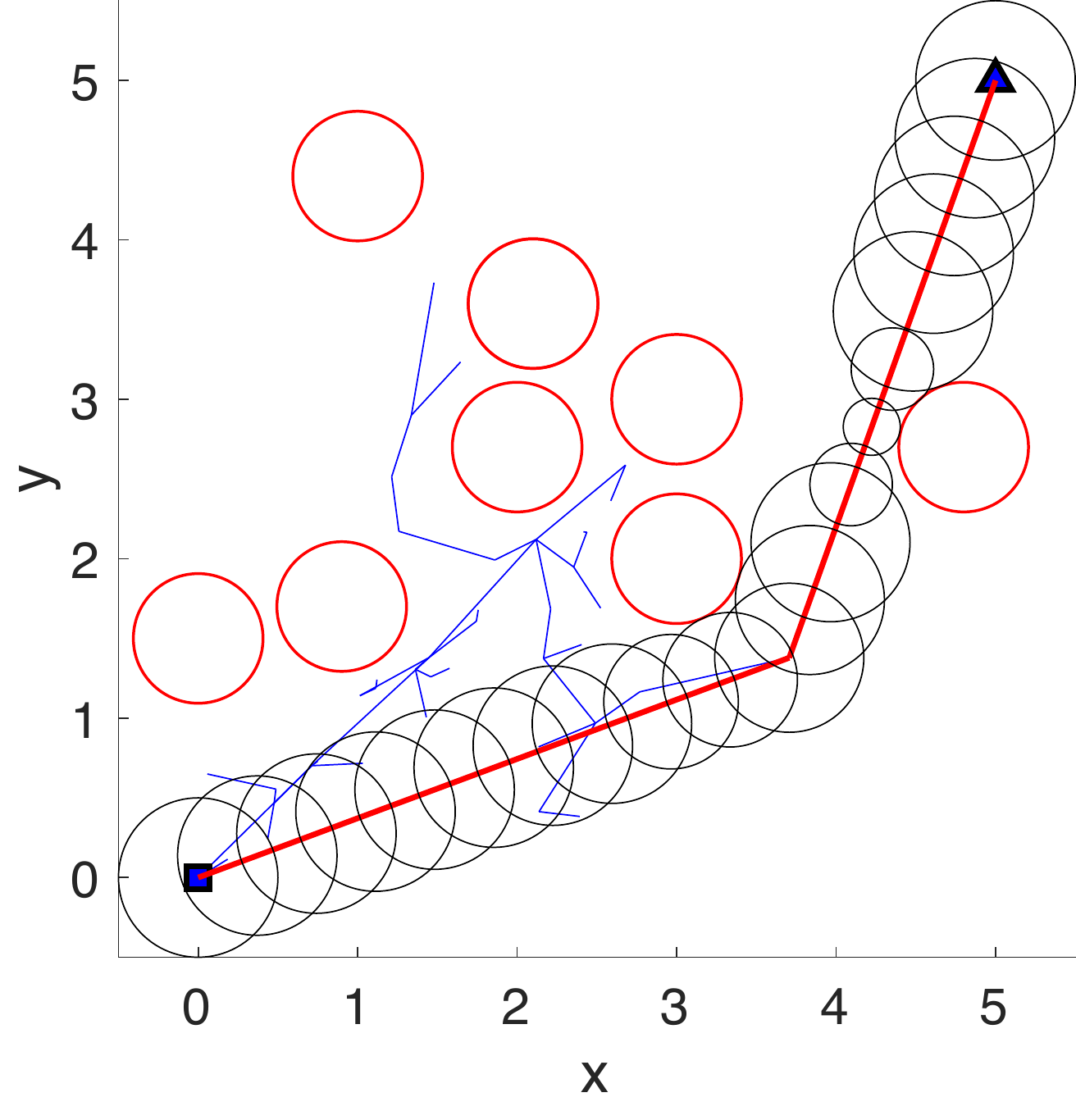}
	\caption{{Top: Example \ref{sec:exp_d_1}. Risk bounded trajectory between the start (square) and goal (triangle) points  obtained via the RRT-SOS algorithm in the static uncertain cluttered environment. Red circles are risk contours of obstacles.
	The blue line segments are the edges of RRT tree. The red line segments are the final path.}
    Bottom: Example \ref{sec:exp_d_2}. Risk bounded tube (black circles) between the start and goal points. The size of the tube at any point along the trajectory is at least 0.1. 
	} 
  \label{fig_exa4_1}
\end{figure}

\begin{figure}
    \centering
    \includegraphics[scale=0.45]{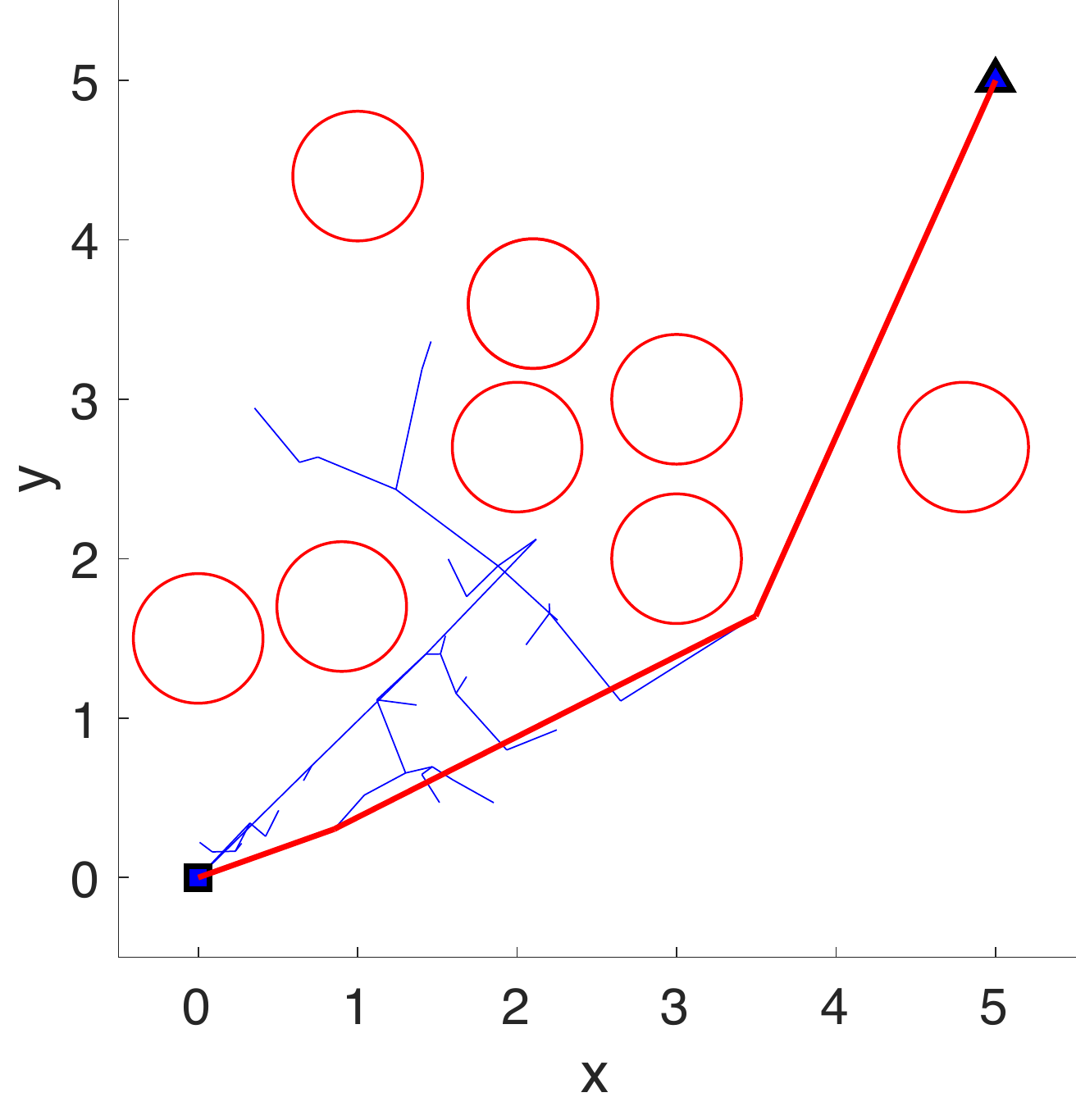}
    \caption{CC-RRT: The planned trajectory is represented by the red line segments. The blue square represents the expected initial position, and the blue triangle represents the expected goal position. The initial position of the system has Gaussian distribution with standard deviation 0.05. The red circles are obstacles, and the blue line segments are the RRT tree. The planned trajectory has bounded risk of 0.1.}
    \label{fig:thesis_chp5_cmp_ccrrt}
\end{figure}

\begin{figure*}
    \centering
    \includegraphics[scale=0.75]{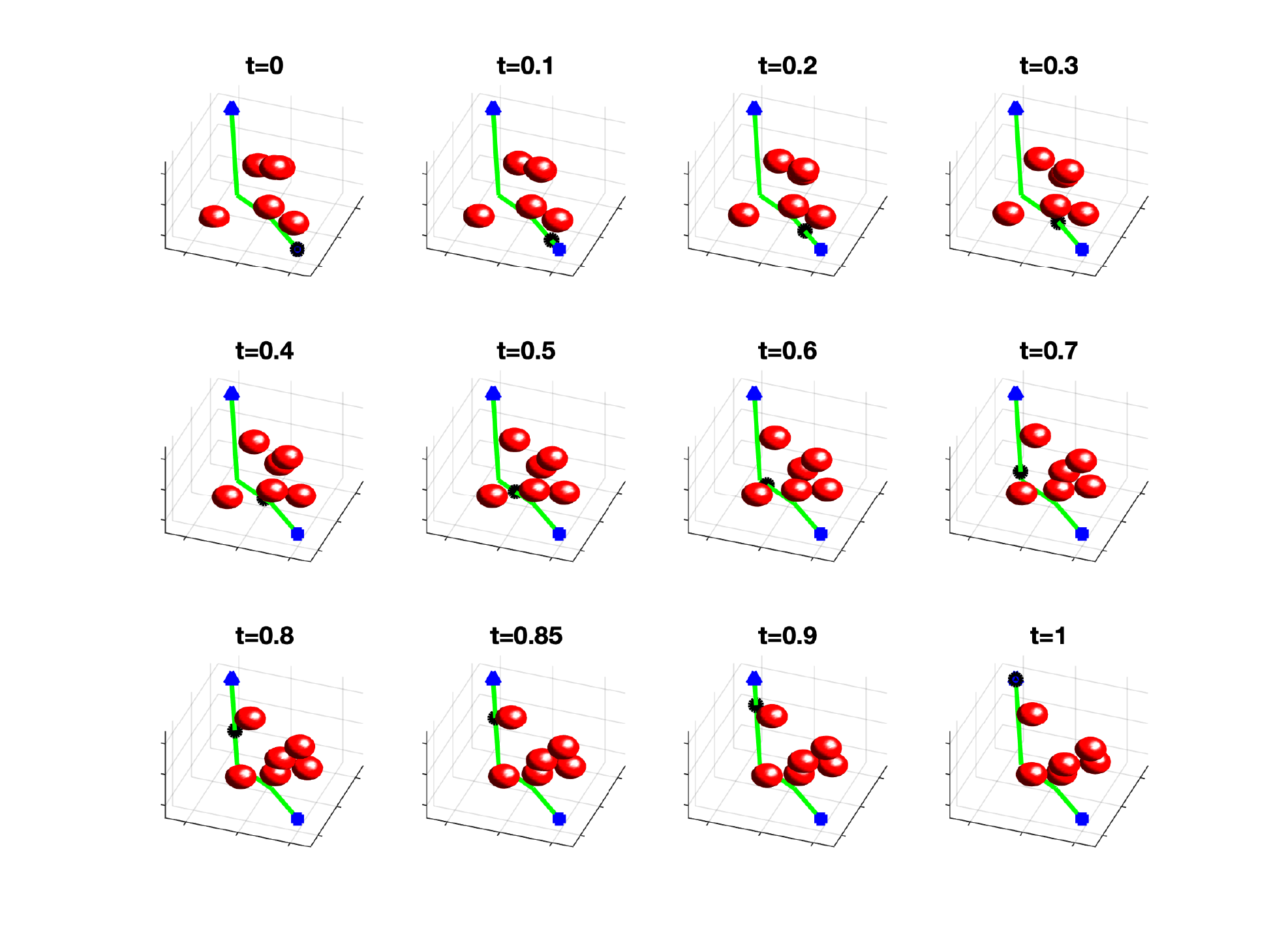}
	\caption{{Example \ref{sec:exp_d_3}: Risk bounded trajectory between the start (square) and goal (triangle) points  obtained via the RRT-SOS algorithm in the dynamic uncertain cluttered environment. Red balls are risk contours of obstacles.}}
  \label{fig_exa4_2}
\end{figure*}
\begin{figure*}
    \centering
    \includegraphics[scale=0.75]{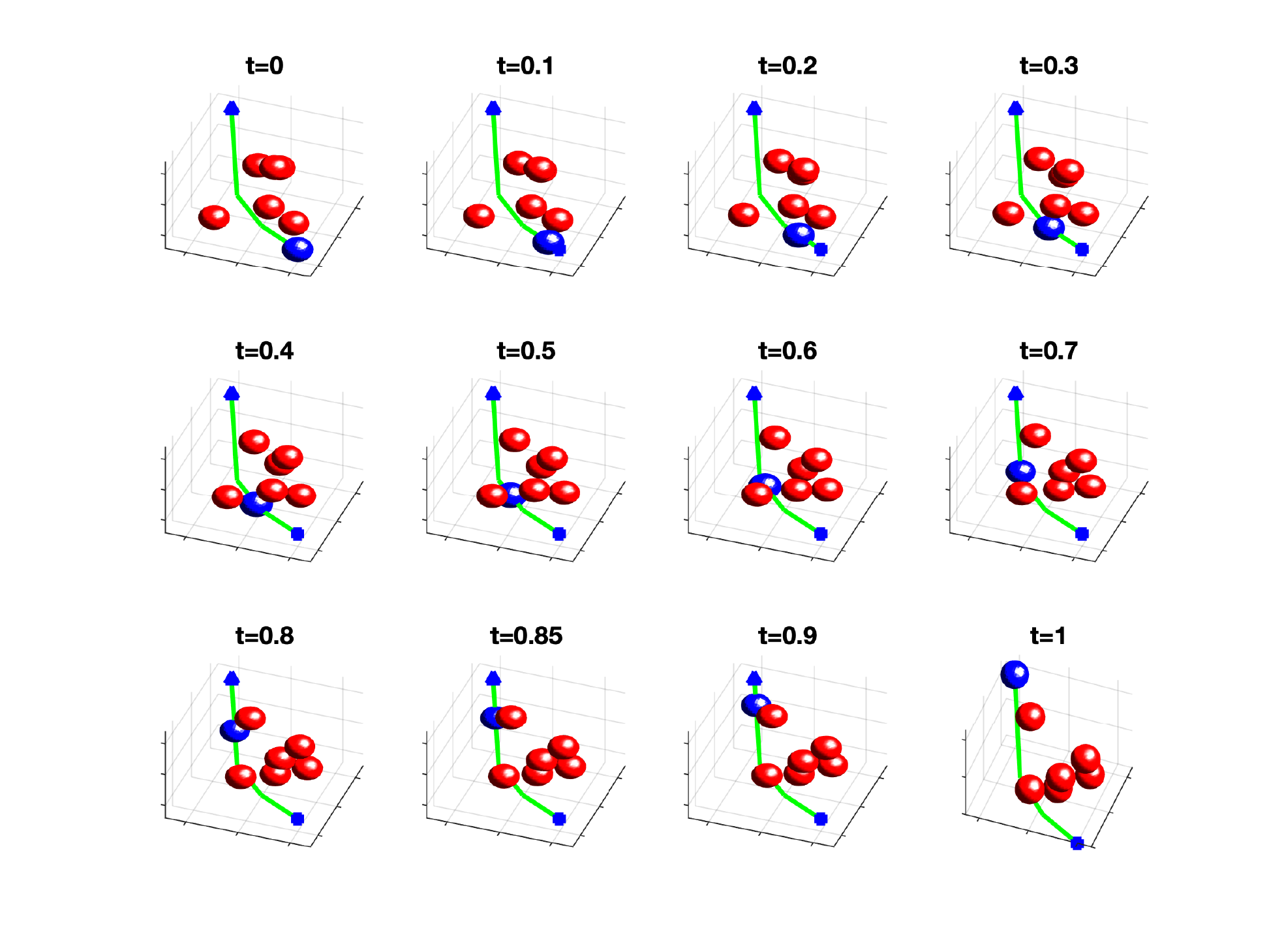}
	\caption{{Example \ref{sec:exp_d_4}: Risk bounded tube (blue balls) whose size is at least 0.1 anywhere between the start (square) and goal (triangle) points in the dynamic uncertain cluttered environment.}}
  \label{fig_exa4_2_2}
\end{figure*}

\subsubsection{Risk Bounded Tube Design} \label{sec:exp_c2}
In this example, we build a max-sized tube directly on top of the piecewise-linear trajectory obtained by RRT-SOS in \ref{sec:exp_c1}. The entire trajectory consists of four line segments. For each line segment except the last one, we build a constant sized tube around the trajectory. For the last line segment, part of the trajectory is very close to the risk contours. If we build a constant tube on this segment, the tube size would be very small on the entire line segment and hence conservative. So we subdivide the line segment into 3 pieces, and build a constant tube around each piece. Figure \ref{fig_exa3_2_2} shows the tube and the underlying trajectory. This is an example where we build a tube after planning a trajectory, and the tube size can be quite small when the trajectory is very close to risk contours.   

\subsection{Planning in Cluttered Uncertain Environments}\label{sec:exp_d}
 In this example, we look for risk bounded trajectories and risk bounded tubes in cluttered static and dynamic uncertain environments.
 
\subsubsection{2D Static Environment Trajectory Planning}\label{sec:exp_d_1}
In this scenario, we use the RRT-SOS algorithm to obtain a risk bounded trajectory with $\Delta=0.1$ between the start and goal points, $\mathbf{x}(0)=(0,0)$, $\mathbf{x}(1)=(5,5)$, in the presence of circle-shaped obstacles with uncertain position as shown in Figure \ref{fig_exa4_1}. The position of the obstacles is subjected to an additive zero mean Gaussian noise with 0.001 variance. 
The run-time to obtain the feasible and optimal trajectories are roughly 27 and 92 seconds, respectively. 

\subsubsection{2D Static Environment Tube Design}\label{sec:exp_d_2}
We look for tubes with minimum size $0.1$ and maximum size $0.5$ between the start and goal points in the 2D static environment described in \ref{sec:exp_d_1}. The trajectory planned in \ref{sec:exp_d_1} does not meet the requirement, because part of the trajectory is almost tangent to the risk contour. A new underlying trajectory is planned together with the tube. The tube consists of several pieces, each having either constant size or quadratic time-varying size. 

\subsubsection{Comparison with Chance-Constrained RRT Method}
We compare our tube design method, for example Method (2), with Chance-Constrained RRT (CC-RRT) method proposed in \cite{luders2010chance}. 
In their setting, the system dynamics is linear and has Gaussian noise. 
The initial position of the system has Gaussian distribution.
The obstacles are convex polyhedra. 
We modify the CC-RRT method to exclude system dynamics. 
We assume the system has an initial Gaussian distribution, and the distribution does not change over time.

We plan a trajectory using CC-RRT in a clustered environment as shown in \Cref{fig:thesis_chp5_cmp_ccrrt}. 
The planned trajectory is represented by the red line segments.
The initial position of the system has Gaussian distribution with standard deviation 0.05.
The red circles are obstacles, and the blue line segments are the RRT tree.
The planned trajectory has bounded risk of 0.1.

Our method has three advantages over this simplified version of CC-RRT, which does not consider system dynamics: i) Our method can ensure the safety of the trajectory and the tubes over the entire continuous time interval, while CC-RRT can only check certain discrete points on the planning horizon. ii) Our method can deal with more general probabilistic distributions, while CC-RRT only works with Gaussian distribution. iii) Our method can deal with more general obstacles, and does not assume obstacles are convex. 

\subsubsection{3D Dynamic Environment Trajectory Planning} \label{sec:exp_d_3}
In this scenario, we use the RRT-SOS algorithm to obtain a risk bounded trajectory with $\Delta=0.1$ between the start and goal points, $\mathbf{x}(0)=(-1,-1,-1)$, $\mathbf{x}(1)=(1,1,1)$, in the presence of moving sphere-shaped obstacles with uncertain radius and uncertain trajectories as shown in Figure \ref{fig_exa4_2}. Radius of the obstacles has a uniform distribution over $[0.1,0.2]$. Also, trajectories of the obstacles are subjected to an additive zero mean Gaussian noise with 0.001 variance. The run-time to obtain the risk bounded trajectory is roughly 29 seconds.

\subsubsection{3D Dynamic Environment Tube Design} \label{sec:exp_d_4}
We look for tubes with minimum size 0.1 between the start and goal points in the 3D dynamic environment described in \ref{sec:exp_d_3}. The trajectory planned in \ref{sec:exp_d_3} does not meet the requirement. A new underlying trajectory is planned together with the tube as shown in Figure \ref{fig_exa4_2_2}. The tube consists of three parts, each having constant size. 

\subsection{Discussion}\label{sec_discuss}

The provided risk bounded algorithms not only are faster than Monte Carlo-based risk bounded RRT algorithms (illustrative example 3), but also provide continuous-time trajectories with guaranteed bounded risk. In addition, compared with the RRT-SOS algorithm, the time-varying SOS optimization is generally faster. The run-time of the time-varying SOS optimization is a function of the order of polynomials of risk contours, the number of the linear pieces of risk bounded trajectories, and the number of iterations of the heuristic algorithm. We use the heuristic algorithm introduced by \cite{SOS_time2} that trades off theoretical guarantees for more efficiency to avoid large scale time-varying SOS optimization. Hence, the heuristic algorithm may fail to obtain a feasible trajectory as observed in Experiment D. To ensure safety, we verify the trajectory returned by the heuristic algorithm using the SOS condition in (\ref{SOS_Cond}). In the time-varying SOS optimization, one can initially start with a small number of iterations and linear pieces and then increase the parameters if the obtained trajectory is infeasible. On the other hand, the RRT-SOS algorithm is more robust and always returns a feasible trajectory. The provided SOS-based RRT algorithm uses a naive tree search algorithm to look for the risk bounded trajectories. One can improve the performance and run-time by considering efficient sampling-based motion planning algorithms and high-order polynomial trajectories between the samples.

Our algorithms have several limitations.
\begin{enumerate}
    \item Regarding the analytical method for obtaining the risk contours, the problem of how good the approximation is is equivalent to the problem of how tight Cantelli's inequality \eqref{cheby2} is, ``tight'' in the sense that the gap between the two sides of the inequality is small.
    In general, the question is how tight the concentration inequalities, such as Markov's inequality, Chebyshev's inequality, and Cantelli's inequality, are.
    There are probability distributions where the equality holds in concentration inequalities.
    However, for a general probability distribution, we do not have any quantitative results on the gap between the two sides of the concentration inequality. 
    We only know that empirically most of the time the concentration inequality works well. 
    \item We assume that the future trajectories and the uncertainty of the obstacles are known. 
    As explained in \Cref{sec_prob}, this is because there are plenty of techniques for predicting future trajectories of obstacles with uncertainty, especially in the field of autonomous driving. 
    The prediction of future trajectories is not guaranteed to be accurate. 
    To mitigate the effect of inaccurate predictions, in the real world implementation, such as on an autonomous driving vehicle, the prediction and the planning should be carried out every few milliseconds, much as in the model predictive control (MPC) fashion, and the system is updated with the latest environment information for re-prediction and re-planning.
    \item We assume that the moments of the uncertainty can be calculated and are finite. Our methods cannot deal with probability distributions that do not have finite moments, such as Cauchy distribution. 
    For probability distributions that cannot be calculated analytically, we need to use sampling to approximate the moments.   
    \item Our algorithms do not consider system dynamics. For the planner that considers stochastic nonlinear dynamics, please refer to \cite{han2022non}. 
\end{enumerate}

\section{Conclusion}\label{sec_concl}
In this paper, we provided continuous-time trajectory planning algorithms to obtain risk bounded polynomial trajectories in uncertain nonconvex environments that contain static and dynamic obstacles with probabilistic location, size, and geometry. The provided algorithms  leverage  the  notion  of  risk  contours to transform the probabilistic trajectory planning problem into a deterministic planning problem and use convex methods to obtain the continuous-time trajectories with guaranteed bounded risk without  the  need  for  time discretization and uncertainty samples. The provided algorithms are suitable for online and large scale planning problems. 
In addition, we provided continuous-time tube design algorithms that build max-sized tubes around trajectories so that any states deviating from the nominal trajectory but inside the tube are still guaranteed to have bounded risk.
In this work, we do not consider system dynamics. 
For the future work, we will take system dynamics into consideration. 
We will also consider more uncertainties in the system, such as uncertainty in initial positions, and uncertainty in system dynamics. 
We will plan risk bounded trajectories for stochastic nonlinear systems.

\bibliographystyle{SageH}
\bibliography{references} 

\end{document}